\documentclass[times,twocolumn,final,authoryear]{elsarticle}

\usepackage{ycviu}
\usepackage{framed,multirow}
\usepackage{amsmath}
\usepackage{amssymb}
\usepackage{latexsym}
\usepackage{hyperref}
\usepackage{url}
\usepackage[table]{xcolor}
\usepackage{xcolor}
\definecolor{newcolor}{rgb}{.8,.349,.1}
\usepackage{subfigure}

\begin{document}

\thispagestyle{empty}

\clearpage
\thispagestyle{empty}
\ifpreprint
  \vspace*{-1pc}
\fi

\clearpage
\thispagestyle{empty}

\ifpreprint
  \vspace*{-1pc}
\else
\fi






 



\clearpage

\ifpreprint
  \setcounter{page}{1}
\else
  \setcounter{page}{1}
\fi

\begin{frontmatter}

\title{Multi-label Image Classification using Adaptive Graph Convolutional Networks: from a Single Domain to Multiple Domains}

\author[1]{Inder Pal \snm{Singh}\corref{cor1}} 
\cortext[cor1]{Corresponding author: 
  }
\ead{inder.singh@uni.lu}
\author[1,2]{Enjie \snm{Ghorbel}}
\author[1]{Oyebade \snm{Oyedotun}}
\author[1]{Djamila \snm{Aouada}}

\address[1]{Interdisciplinary Centre for Security, Reliability and Trust (SnT), University of Luxembourg, Luxembourg}
\address[2]{Cristal Laboratory, National School of Computer Sciences, University of Manouba, Tunisia}

\received{1 May 2013}
\finalform{10 May 2013}
\accepted{13 May 2013}
\availableonline{15 May 2013}
\communicated{S. Sarkar}

\begin{abstract}
This paper proposes an adaptive graph-based approach for multi-label image classification. Graph-based methods have been largely exploited in the field of multi-label classification, given their ability to model label correlations. Specifically, their effectiveness has been proven not only when considering a single domain but also when taking into account multiple domains. However, the topology of the used graph is not optimal as it is pre-defined heuristically. In addition, consecutive Graph Convolutional Network (GCN) aggregations tend to destroy the feature similarity. To overcome these issues, an architecture for learning the graph connectivity in an end-to-end fashion is introduced. This is done by integrating an attention-based mechanism and a similarity-preserving strategy. The proposed framework is then extended to multiple domains using an adversarial training scheme. Numerous experiments are reported on well-known single-domain and multi-domain benchmarks. The results demonstrate that our approach achieves competitive results in terms of mean Average Precision (mAP) and model size as compared to the state-of-the-art. The code will be made publicly available. 
\end{abstract}


\end{frontmatter}


\section{Introduction}
\label{sec:intro}
Multi-label image classification has been widely investigated by the computer vision community, given its practical relevance in numerous application fields, such as human attribute recognition~\citep{human_attr}, scene classification~\citep{scene1} and deepfake detection~\citep{mmspInder}. In contrast to traditional image classification approaches that associate a single label to an input image, multi-label image classification aims to detect the presence of a set of objects.

\begin{figure}[t]
\centering
\subfigure[]{\includegraphics[width=0.45\textwidth]{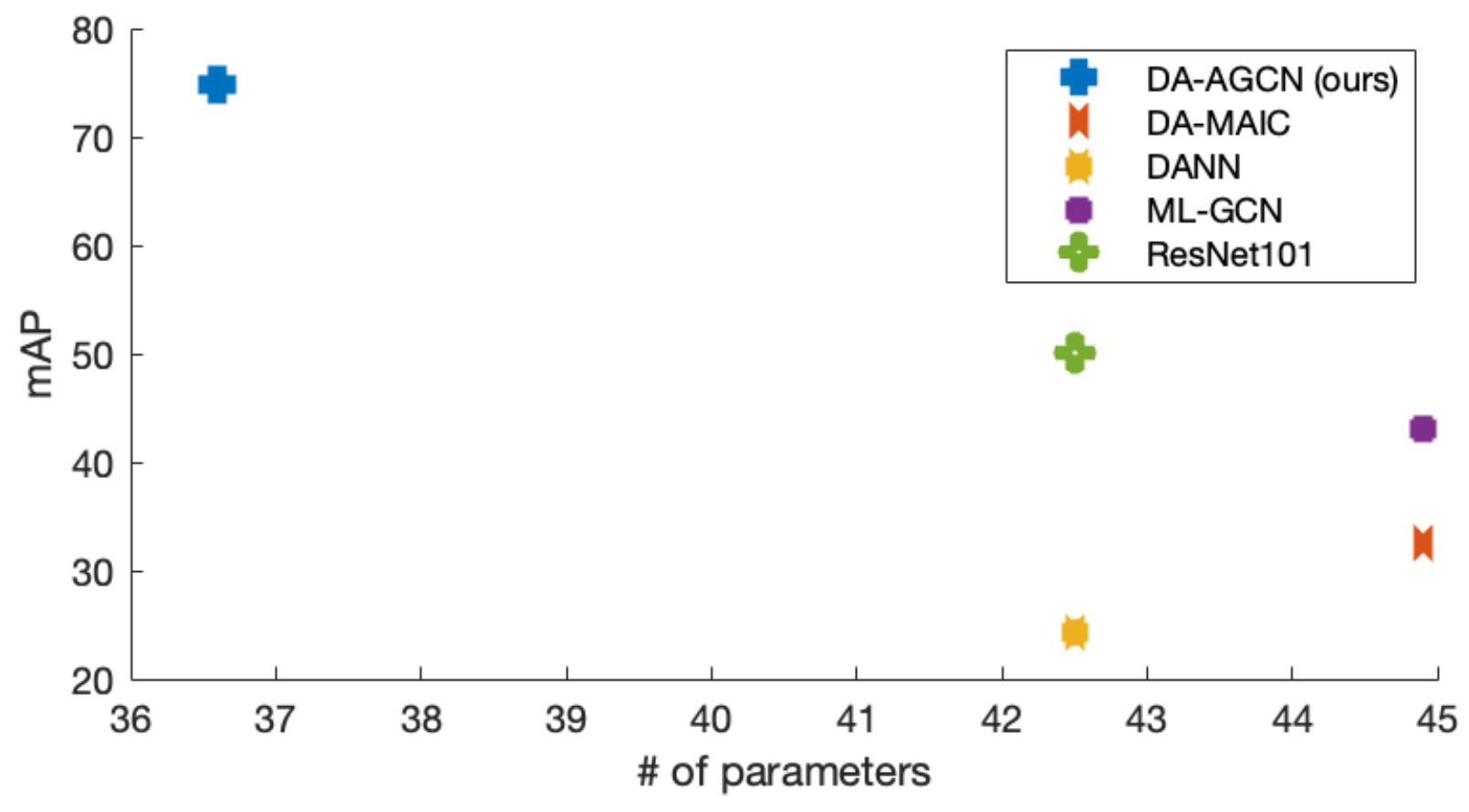}}
\subfigure[]{\includegraphics[width=0.45\textwidth]{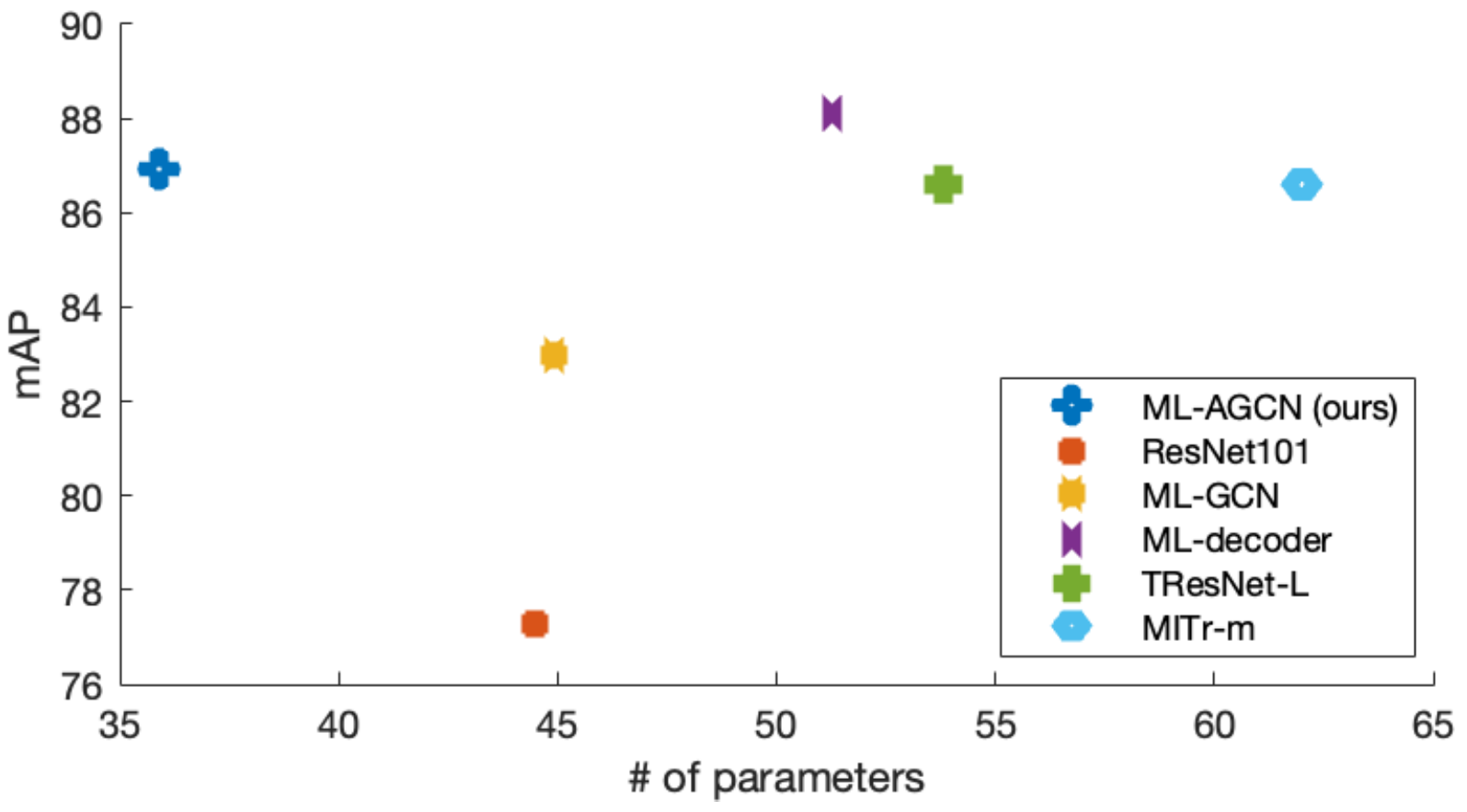}}
    \caption{Comparison of our approach (ML-AGCN) without (top) and with UDA (down) to recent state-of-the-art methods in terms of number of parameters (millions) and mean Average Precision (mAP) on MS-COCO and Clipart $\rightarrow$ VOC. The considered state-of-the art methods are: MlTr-m~\citep{mltr}, TResNet-L~\citep{tresnet}, ML-Decoder~\citep{ml-decoder}, ML-GCN~\citep{ml-gcn}, ResNet101~\citep{resnet}, DA-MAIC~\citep{da-maic}, and DANN~\citep{dann}.}
\label{fig:intro}
\end{figure}

Thanks to the latest progress in deep learning, most of recent methods rely on a single-stream Deep Neural Networks (DNN), including Convolution Neural Networks (CNN)~\citep{vgg, resnet, srn, multi-evidence, mcar, tresnet}, Residual Neural Networks (RNN)~\citep{cnn-rnn} and Transformers~\citep{c-trans,mltr, stmg, vg-500}. Nevertheless, their impressive performance comes at the cost of very large architectures that are unsuitable for memory-constrained environments. As an alternative, another line of research has tried to integrate priors related to label correlations~\citep{ml-gcn, ssgrl, a-gcn, f-gcn, iml-gcn, cfmic, FLNet}. As demonstrated in~\citep{iml-gcn}, such an approach contributes to improving scalability. In other words, fewer parameters are required to achieve comparable performance with traditional DNNs.

Graph-based approaches~\citep{ml-gcn,iml-gcn}, are among the most popular multi-label classification methods that aim at modeling label correlations. In addition to a standard DNN that extracts image features, a second stream based on a Graph Convolutional Network (GCN) is used for generating inter-dependent label classifiers. In particular, the input graph is used to model label dependencies. Each node represents a label, and each edge is characterized by the probability of co-occurrence of a label pair.  Graph-based methods have also been shown to be successful in the challenging context of cross-domain multi-label classification~\citep{da-maic}.   Standard multi-label image classification methods such as~\citep{ml-gcn,iml-gcn}  usually assume that unseen images and training data are drawn from the same distribution, i.e. the same domain, hence ignoring a possible domain shift problem~\citep{da-maic}. This leads, therefore, to poor generalization capabilities under cross-domain settings. UDA is a plausible solution to overcome this challenge without relying on costly annotation efforts~\citep{uda}. UDA aims at learning domain invariant features to bridge the gap between a \textit{source} domain and a \textit{target} domain without access to the associated labels. In particular, one of the most successful UDA approaches for multi-label classification, namely DA-MAIC~\citep{da-maic}, leverages graph representations to model label inter-dependencies and couple it with an adversarial training approach~\citep{dann}. While most existing multi-label classification methods can be extended to the cross-domain setting by just adding a simple discriminator, they mostly require a high number of parameters to work effectively. Hence, the use of a graph-based method for modeling label correlation is an interesting way for obtaining compact yet effective models, as highlighted in ~\citep{da-maic}.  This allows achieving a good compromise between performance and compactness under the cross-domain setting.


Despite their usefulness in both single and cross-domain settings, graph-based methods~\citep{ml-gcn, ssgrl, a-gcn, f-gcn, iml-gcn, cfmic, FLNet, da-maic} are unfortunately subject to three major limitations, namely: (1) The graph structure is heuristically defined. In particular, it is computed based on the co-occurrence of labels in the training data. Hence, this topology might not be ideal for the specific task of multi-label image classification; (2) A threshold is empirically fixed for discarding edges with a low co-occurrence probability. This means that infrequent co-occurrences are assumed to be noisy. Although this might be true in many cases, assuming that any rare event corresponds to noise does not always hold; and (3) it has been proven in~\citep{nodesim} that successive aggregation operations in the GCN usually dissipate the node similarity in the original feature space, hence potentially leading to a decrease in terms of performance. 


Herein, we posit that by integrating adequate mechanisms in graph-based approaches for addressing the aforementioned issues, it should be possible to reduce the network size even more while achieving competitive performance in both single-domain and cross-domain settings.

In this paper, we propose an adaptive graph-based multi-label classification method called Multi-Label Adaptive Graph Convolutional Network (ML-AGCN) for both contexts, single domain and across domains. Our idea consists in: (1) learning two additional adjacency matrices in an end-to-end manner instead of solely relying on a heuristically defined graph topology. Note that no threshold is applied, avoiding the loss of weak yet relevant connections. In particular, the first learned graph topology computes the importance of each node pair. This is carried out by employing an attention mechanism similar to Graph Attention Networks (GAT)~\citep{gat}. {Nevertheless, even though learned, the latter does not ensure the conservation of feature similarity}. Hence, the second graph structure is built based on the similarity between node features and overcomes the information loss happening through successive convolutions; (2) integrating the proposed adaptive graph-based architecture in an adversarial domain adaptation framework for aligning a labeled source domain to an unlabeled target domain. As shown in Fig.~\ref{fig:intro}, the results suggest that our method is competitive with respect to the state-of-the-art in terms of both mean Average Precision (mAP) and network size under the single domain and cross-domain settings.

This paper is an extended version of~\citep{ml-agcn}. In comparison to~\citep{ml-agcn}, the main contributions of this article are given below:

\begin{enumerate}
     \item More extensive experiments showing the relevance of the proposed architecture called Multi-Label Adaptive Graph Convolutional Network (ML-AGCN) for multi-label image classification.

    \item An adversarial Domain Adaptation approach integrating an Adaptive Graph Convolutional Network (DA-AGCN) for multi-label image classification.

    \item A deep qualitative and quantitative experimental analysis of the proposed domain adaptation method with respect to the state-of-the-art.

\end{enumerate}    

The remainder of this paper is organized as follows. Section~\ref{sec:related_works} reviews the state-of-the-art on multi-label image classification and domain adaptation for multi-label classification.  Section~\ref{sec:background} formulates the problem of graph-based multi-label image classification and its applicability to different domains. In Section~\ref{sec:method_1} and Section~\ref{sec:method_2}, the proposed method is detailed.  Section~\ref{sec:exp} presents the experimental results and analysis. Finally, Section~\ref{sec:conclusion} concludes this work and highlights interesting future directions.

\section{Related works}
\label{sec:related_works}
 In this section, we start by presenting the state-of-the-art of multi-label image classification. Then, we focus on reviewing existing domain adaptation approaches for multi-label image classification. 

\subsection{Multi-label Image Classification}
\label{sec:related_works_mlic}
This subsection presents various works proposed for Multi-Label Image Classification (MLIC) within a single-domain context. We categorize these methodologies into two main groups based on their intuition and network architectures.

\subsubsection{Single-stream deep neural networks}
As discussed in Section~\ref{sec:intro}, most multi-label classification methods~\citep{cnn-svm, hcp, c-trans, mltr, stmg, tresnet, asl} employ a single-stream DNN. More specifically, they mainly take inspiration from successful architectures proposed in the context of single-label image classification such as CNN, RNN, and transformers. For example, \cite{cnn-svm} have used a pre-trained OVERFeat~\citep{overfeat} model and have adapted it to multi-label image classification. \cite{hcp} have leveraged the prediction of multiple CNN architectures pretrained on ImageNet~\citep{imagenet} such as AlexNet~\citep{alexnet} and VGG-16~\citep{vgg}.
Ridnik et al.~\citep{asl} have employed TResNet~\citep{tresnet} using a novel loss called Asymmetric Loss (ASL) that focuses more on positive labels than negative ones. TResNet introduced in~\citep{tresnet} is based on a ResNet architecture with a series of modifications for optimizing the GPU network capabilities while maintaining the performance.

Recently, ~\cite{c-trans} attempted to leverage transformers for modeling complex dependencies among visual features and labels. Similarly, MlTr~\citep{mltr} combines the pixel attention and the attention among image patches to better excavate the transformer’s activity in multi-label image classification. More recently, ML-Decoder~\citep{ml-decoder} proposed a transformer-based classification head instead of the standard Global Average Pooling (GAP), for improving the generalization capability. On the other hand, another recent work introduced a novel attention module called Interventional Dual Attention (IDA)~\citep{ida} that aims to mitigate contextual bias in visual recognition through multiple sampling interventions.

\subsubsection{Multi-stream deep neural networks}

Going deeper into the network enables the model to learn more abstract features from the image. However, this comes at the cost of high memory requirements. Moreover, the aforementioned single-stream methods do not explicitly model the relationship between labels which can be an important semantic element to consider.

In order to incorporate the information of label correlations, a second class of methods have used a second subnetwork in addition to the main backbone. For instance, \cite{srn} have introduced a Spatial Regularization Net (SRN) to learn the underlying relationships between labels by generating label-wise attention maps.
Similarly, \cite{msrn} have employed image representations from intermediate convolutional layers to model both local and global label semantics.

Graphs can also be an interesting way for modeling label correlations~\citep{ml-gcn, iml-gcn, kssnet, stmg}, while keeping the network size reasonable. 
ML-GCN~\citep{ml-gcn} was among the pioneering works that utilized a Graph Convolutional Network (GCN) to learn interdependent label-wise classifiers and combine them with a standard DNN that learns discriminative image features. 
Similarly, IML-GCN~\citep{iml-gcn} makes use of a similar GCN-CNN architecture. However, they suggest employing image-based embeddings as node features of the input graph as a replacement to the glove-based word embeddings~\citep{glove} used in~\citep{ml-gcn}. Indeed, GLOVE embeddings might be inappropriate for MLIC since they have has been initially proposed for representing words in the field of Natural Language Processing (NLP), while images are intrinsically different.

As discussed in Section~\ref{sec:intro}, despite their proven performance in terms of both precision and network size, these graph-based approaches including ML-GCN and IML-GCN suffer from some weaknesses. First, the input graph is predefined based on label co-occurrences in the training set. As a result, the heuristically defined topology labels might be sub-optimal for MLIC. Second, a threshold is set empirically in order to ignore weak edges. This induces that rare co-occurrences are presumed to be noisy. Even though this assumption often holds, some infrequent events might occur in a real-life scenario. Lastly,  as discussed in~\citep{nodesim}, the similarity of node features through multiple GCN layers might fade, potentially resulting in a performance decrease. The proposed ML-AGCN aims at solving these issues by adaptively learning the graph topology in an end-to-end manner, while leveraging an adequate mechanism for preserving the feature node similarity.

\subsection{Unsupervised Domain Adaptation for Multi-label Image Classification}
Over the last years, DL methods have achieved a remarkable progress in the field of computer vision and pattern recognition. However, the effectiveness of DL approaches heavily depends on the availability of a large amount of annotated data. For mitigating the huge cost caused by data annotation, the field of unsupervised domain adaptation~\citep{etd, mdd, mmd, dann, cada} has been widely investigated over the last decade. It aims at making use of an existing labeled dataset from a related domain called \textit{source domain} to
enhance the model performance on a domain of interest termed \textit{target domain}, for which only unlabeled data are provided.

Unsupervised domain adaptation methods can be separated into two main categories. The first one~\citep{etd,mdd,mmd} aims at explicitly reducing the domain gap by minimizing statistical discrepancy measures between the two domains. Alternatively, the second class of methods implicitly minimizes this domain gap by adopting an adversarial training approach~\citep{dann,cada}. The main idea consists in using a domain classifier to play a min-max two-player game with the feature generator. This strategy is designed to enforce the generation of domain-invariant features that are sufficiently discriminative.

Nevertheless, most existing techniques focus on the task of single-label image classification. In fact, very few papers have considered domain adaptation for multi-label image classification~\citep{ml-anet, chest, ml-gcn},\citep{dda-mlic}.   

Among these rare references, we can mention ML-ANet~\citep{ml-anet}, which explicitly minimizes the domain gap by optimizing multi-kernels maximum mean discrepancies (MK-MMD) in a Reproducing Kernel Hilbert Space (RKHS). 
More recently, an adversarial approach has been adopted in~\citep{chest} where a condition-based domain discriminator similar to conditional-GANs~\citep{c-gan} has been employed. However, similar to the first category of methods for traditional multi-label image classification, these two approaches~\citep{ml-anet, chest} neglect the important information of label dependencies. 

A graph-based approach called DA-MAIC has been then proposed as an alternative~\citep{da-maic}. As in ML-GCN~\citep{ml-gcn}, they have proposed to build a graph for modeling label correlations based on label co-occurrences. Additionally, to reduce the domain shift between the source and target domains, they have used a domain classifier that is trained in an adversarial manner. Unfortunately, DA-MAIC is impacted by the same drawbacks affecting ML-GCN~\citep{ml-gcn} as detailed in Section~\ref{sec:intro} and~\ref{sec:related_works_mlic}. More details regarding these issues are given in the next section. 

Inspired by~\citep{daln}, a very recent work~\citep{dda-mlic} called DDA-MLIC redefining the adversarial loss based on the multi-label classifier has been proposed, hence eliminating the need for an additional discriminator. Although this approach has shown promising results, it addresses the problem of domain adaptation from a different perspective. Indeed, the present paper aims to propose a suitable multi-label classification mechanism that could be beneficial for both single-domain and cross-domain settings while DDA-MLIC investigates a more adequate adversarial strategy.  Therefore, the two methods (DDA-MLIC and ours) tackle two different yet complementary aspects of UDA for multi-label classification.

\section{Background and Problem Formulation}
\label{sec:background}

In this section, Graph Convolutional Networks (GCN) are first reviewed. Then, the problem of multi-label image classification in both single and cross-domain settings is formulated.   

\subsection{Background: Graph Convolutional Networks (GCN)}
CNNs are defined on a regular grid and are, therefore, not directly applicable to non-Euclidean structures such as graphs. GCN~\citep{graph} have been proposed as the generalization of traditional CNN to graphs. They have been very successful in numerous computer vision applications, including human pose estimation~\citep{human_pose} and human action recognition~\citep{human_action,fgkostas}.
Let us denote a graph by $G = (V,E,\mathbf{F})$. The set $V=\{v_1, v_2, ..., v_N\}$ is formed by $N$ nodes, while $E=\{e_1, e_2, ..., e_M\}$ refers to the set of $M$ edges connecting the nodes. Finally, $\mathbf{F}=\{\mathbf f_1, \mathbf f_2,..., \mathbf f_N\}$ represents the node features such that $\mathbf{f}_i \in \mathbb{R}^d$ corresponds to the features of the node $v_i$.

Let us assume that $\mathbf{F}^l$ is the input node features of the $l^{\text{th}}$ layer and $\mathbf A\in \mathbb{R}^{N \times N}$ the adjacency matrix of the graph $\mathcal{G}$. Each GCN layer can be seen as a non-linear function $h(.)$ that computes the node features  $\mathbf F^{l+1} \in \mathbb{R}^{N \times d'}$ of the $(l+1)^{\text{th}}$ layer as follows,

\begin{equation}
\label{eq:gcn-conv}
    \mathbf F^{l+1} = h(\mathbf A \mathbf F^l \mathbf W^l),
\end{equation}
where $\mathbf W^l \in \mathbb{R}^{d\times d'}$ is the weight matrix. Note that $\mathbf A$ is normalized before using Eq.~\eqref{eq:gcn-conv}.

\subsection{Problem Formulation}
\subsubsection{Graph-based multi-label image classification}
\label{problem-form-graph}
The goal of multi-label image classification is to predict the presence or absence of a set of objects $\mathcal O=\{1,2,...,N \}$ in a given image $\mathbf I$. This can be done by learning a function $f$ such that,

\begin{equation}
    \begin{split}
        f \colon & \mathbb{R}^{ w \times h} \to [\![ 0,1 ]\!]^N\\
  & \mathbf I \mapsto \mathbf{y}=[y_i]_{i\in \mathcal{O}},
    \end{split}
\end{equation}

where $w$ and $h$ define the pixel-wise width and height of the image, respectively, and $y_i=1$ indicates the presence of the label $i$ in $\mathbf I$, in contrast to $y_i=0$.
Graph-based multi-label image classification methods such as ML-GCN~\citep{ml-gcn} and IML-GCN~\citep{iml-gcn} usually involve two subnetworks: (1) A feature generator denoted by $f_{g}$, and (2) an estimator of $N$ inter-dependent binary classifiers denoted by $ f_{c}$. The generator mostly corresponds to an out-off-the-shelf CNN network which produces a $d_f$-dimensional image feature representation as described below,

\begin{equation}
\begin{split}
       f_{g} \colon & \mathbb{R}^{ w \times h} \to \mathbb{R}^{d_f}\\
  & \mathbf I \mapsto \mathbf X
\end{split}.
\end{equation}
For instance, ML-GCN integrates a ResNet101~\citep{resnet}, while IML-GCN makes use of TResNet-M~\citep{tresnet}.   

On the other hand, the second subnetwork $f_{c}$ is a GCN formed by $L$ layers which takes a fixed graph $G=(V, {E}, \mathbf{F})$ as input where $\text{card}(V)=N$. In fact, each node $v_i \in V$ refers to the label $i\in \{1,2,...,N\}$ and each $\mathbf{f}_i \in \mathbf F$ corresponds to its associated label embedding.  The adjacency matrix $\mathbf A \in \mathbb{R}^{N \times N}$ of $G$ is usually pre-computed based on the co-occurrence probabilities that are estimated over the training set~\citep{iml-gcn, ml-gcn}.   In addition, rare co-occurrences are discarded as they are assumed to be noisy. In other words, given an empirically fixed threshold $\tau$,

\begin{equation}
\label{eq:threshold}
    \mathbf{A}_{ij} = \left\{\begin{array}{l}
0, \text { if } p_{ij} < \tau, \\
1, \text { if } p_{ij} \geq \tau
\end{array}\right.,
\end{equation}
with $p_{ij}=p(I_j|I_i)$ being the probability of co-occurrence of label $j$ and label $i$ in the same image.

Given $\mathbf{F}^l \in \mathbb R^{d_l \times L}$ as the input node features of the $l^{\text{th}}$ layer, the  input features $\mathbf{F}^{l+1}$ of the $(l+1)^{\text{th}}$ are therefore computed by following Eq.~\eqref{eq:gcn-conv}.

Finally, the vertex features produced by the last GCN layer, i.e., $\mathbf{F}^{L} \in \mathbb{R}^{N \times d_f}$, form the $N$ inter-dependent classifiers. 

In summary,  $f_{c}$ can be defined as follows,

\begin{equation}
\begin{split}
   f_{c} \colon & \mathcal{G}(N) \to \mathbb{R}^{d_f \times N} \\
  & G \mapsto \mathbf F^L = [F^L_1,...,F^L_N],
\end{split}
\end{equation}

\noindent
where $\mathcal{G}(N)$ represents the set of graphs with $N$ nodes, and $\mathbf F_i^L$ for $i \in \{1,...,N\}$ is the generated inter-dependent binary classifier associated to the label $i$. 

Hence, $f$, which returns the final prediction, can be defined as follows,
\begin{equation}
    \begin{split}
        f(\mathbf I)
    &= \text{sig} (f_{c}(G)f_{g}(\mathbf I)^T) \\
    &= \text{sig} (\mathbf F^L \mathbf X^T),
    \end{split}
    \label{eq_pred}
\end{equation}
where $\text{sig} (x) = \frac{1}{1+e^{-x}}$ is the sigmoid activation function.

However, as explained in Section~\ref{sec:intro}, three main limitations can be noted in graph-based methods: (1) the computation of the adjacency matrix $\mathbf A$ is made heuristically and is decoupled from the training process; (2) a threshold $\tau$ is empirically fixed for completely ignoring rare co-occurrences; and (3) as shown in~\citep{nodesim}, aggregating successively node features in a graph may induce the loss of the similarity/dissimilarity information present in the initial feature space. 

 \subsubsection{Graph-based unsupervised domain adaptation for multi-label image classification}

Let us consider a source dataset for multi-label image classification $\{{(\mathbf{I}_s^k, \mathbf y_s^k)}\}_{k=1}^{n_s}$ drawn from a distribution $\mathcal{D}_s$ and a similar unlabelled target dataset $ \{{(\mathbf{I}^k_t)}\}_{k=1}^{n_t}$  drawn from a different distribution $\mathcal{D}_t$. The matrix $\mathbf I_s^k\in \mathbb R^{w\times h}$ refers to the $k^{\text{th}}$ image sample of $\mathcal{D}_s$ and $\mathbf y_s^k \in [\![ 0,1 ]\!]^N $ is its associated label vector, while $\mathbf I_s^t\in \mathbb R^{w\times h}$ denotes the $k^{\text{th}}$ image sample of $\mathcal{D}_t$. The variables $n_s$ and $n_t$ refer respectively to the total number of samples in $\mathcal{D}_s$ and $\mathcal{D}_s$.  The goal of unsupervised domain adaptation is to train a model using labeled source domain data sampled from $\mathcal{D}_s$ and unlabelled target domain samples from $\mathcal{D}_t$ for making accurate predictions on the target domain. Despite its relevance in real-world applications, few domain adaptation methods have been proposed for multi-label image classification. Among the most recent and accurate approaches, one can mention DA-MAIC~\citep{da-maic}, which also takes advantage of the graph representation for modeling label correlations. It directly extends ML-GCN~\citep{ml-gcn} by integrating an adversarial training for domain adaptation. Consequently, DA-MAIC is subject to the same limitations induced by ML-GCN~\citep{ml-gcn} mentioned in Section~\ref{problem-form-graph}. Our assumption is that by finding the appropriate solutions to the enumerated issues, we can, at the same time, enhance the performance of existing multi-label image classification and domain adaptation for multi-label image classification methods.

\section{Multi-Label Adaptive Graph Convolutional Network (ML-AGCN)}
\label{sec:method_1}
To handle the challenges mentioned in Section~\ref{sec:background}, a novel graph-based approach called Multi-Label Adaptive Graph Convolutional Network (ML-AGCN) is introduced. 

\begin{figure*}[!t]
  \centering
  \centerline{\includegraphics[width=1.0\linewidth]{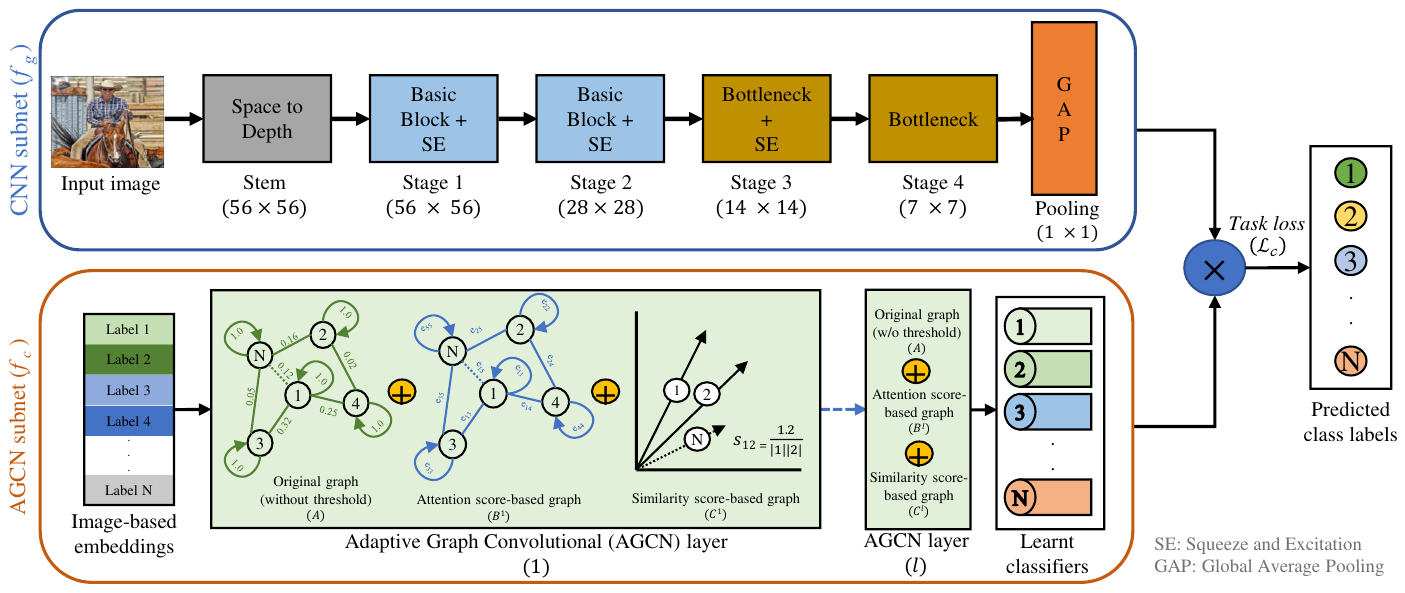}}
\caption{Architecture of ML-AGCN~\citep{ml-agcn}: On the one hand, the CNN subnet learns relevant image features from an input image. On the other hand, the GCN subnet estimates interdependent label classifiers by taking into account one fixed adjacency matrix $\mathbf A$ and two adaptive adjacency matrices $\mathbf B^{(l)}$ and $\mathbf C^{(l)}$. Finally, the classifiers are applied to the CNN features for predicting the labels.}
\label{fig:arch}
\end{figure*}
\subsection{Overview of the Proposed Architecture}
Similar to~\citep{ml-gcn} and \citep{iml-gcn}, a network formed by two subnets is adopted as illustrated in Fig.~\ref{fig:arch}. The first is a CNN that extracts a discriminative representation from a given input image, while the second is a GCN-based network that learns $N$ interdependent classifiers. As in~\citep{iml-gcn}, TResNet-M which represents a small version of TResNet~\citep{tresnet} is employed as a CNN subnetwork. TResNet is a direct extension of ResNet, which fully exploits the GPU capabilities to boost the model efficiency. 
The graph-based subnet, Adaptive Graph Convolutional Network (AGCN), uses the same image embeddings proposed in~\citep{iml-gcn} as feature nodes. Nevertheless, in contrast to~\citep{iml-gcn} and \citep{ml-gcn}, it relies on an end-to-end learned graph topology. More details regarding this subnetwork are provided in Section~\ref{g-subnet}. Similar to~\citep{iml-gcn}, the Asymmetric Loss (ASL)~\citep{asl} denoted by $\mathcal{L}_{c}$ is used for optimizing ML-AGCN such that,

\begin{multline}
    \mathcal{L}_{c} =  \mathbb{E}_{(\mathbf I_s, \mathbf y_s) \sim \mathcal{D}_s} \sum_{i=1}^{N}  y_{s}^{(i)}(1-p^{(i)})^{\gamma_+}\log(p^{(i)}) + \\ (1-y_{s}^{(i)})(p_m^{(i)})^{\gamma_-}\log(1-p_m^{(i)}),
    \label{eq:loss_cls}
\end{multline}
where $\gamma_+$ and $\gamma_-$ are focusing parameters for positive and negative samples, respectively, and $y_s^{(i)}$ and $p^{(i)}$ are the respective ground truth and predicted probability with respect to the label $i$ and $p_m^{(i)}$ is the shifted probability given by $\max (p^{(i)}-m,0)$, where $m$ is a threshold used for reducing the effect of easy negative samples~\citep{asl}. 
\begin{figure}[!t]
\begin{minipage}[b]{.48\linewidth}
  \centering
  \centerline{\includegraphics[width=4cm]{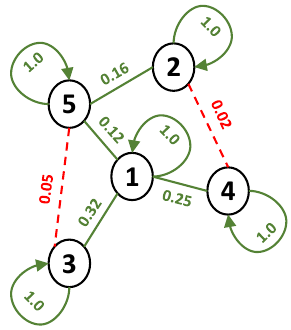}}
  \centerline{(a) Fixed graph $\tau=0.1$}\medskip
\end{minipage}
\begin{minipage}[b]{.48\linewidth}
  \centering
  \centerline{\includegraphics[width=4.0cm]{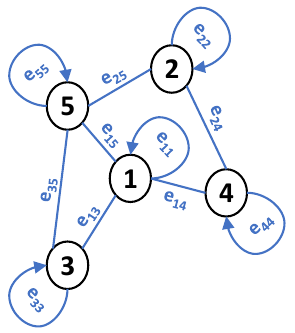}}
  \centerline{(b) Parameterized graph $\tau=0$}\medskip
\end{minipage}
\hfill
\caption{(a) An example of a fixed label graph with a threshold set to $\tau=0.1$~\citep{ml-agcn}. Dashed (red) edges indicate the ignored edges; (b) The proposed parameterized graph topology considering all the edges.}
\label{fig:res}
\end{figure}

\subsection{Graph-based Subnet: Adaptive Graph Convolutional Network (AGCN)}
\label{g-subnet}
Our intuition is that by integrating a suitable mechanism, it should be possible to boost the classification performance and reduce at the same time the size of graph-based methods. Hence, as illustrated in Fig.~\ref{fig:res}, we propose to adaptively learn the graph topology by reformulating Eq.~\eqref{eq:gcn-conv} as below,

\begin{equation}
\label{eq:agcn}
    \mathbf F^{l+1} = \sigma(({\mathbf A}+\mathbf{B}^{(l)}+\mathbf{C}^{(l)})\mathbf F^l \mathbf W^l),
\end{equation}
where $\sigma$ is a LeakyRELU activation function.

Instead of relying solely on the adjacency matrix $\mathbf A$ defined in~\citep{ml-gcn}, two additional parameterized graphs called attention-based and similarity adjacency graphs, respectively denoted by $\mathbf B^{(l)}$ and $\mathbf C^{(l)}$, are defined. In this case, no threshold is applied to $\mathbf A$ for ignoring rare co-occurrences. It is also important to note that $\mathbf A$ is fixed, while $\mathbf B^{(l)}$ and $\mathbf C^{(l)}$ vary from one layer to another. In the following, we detail how $\mathbf B^{(l)}$ and $\mathbf C^{(l)}$ are computed. 

\subsubsection{Attention-based adjacency matrix}

Instead of ignoring rare co-occurrences in $\mathbf A$, the matrix $\mathbf{B}^{(l)}=(b_{ij}^{(l)})_{i,j\in \mathcal{O}}$ is defined based on an attention mechanism where the importance of each edge is quantified. To that aim, inspired by~\citep{gat}, an attention score denoted by $e_{ij}$ is calculated for each pair of vertices $(v_i, v_j)$ as follows,

\begin{equation}
    e_{ij} = \sigma(\mathbf a^{(l)^T} \mathbf (\mathbf{W}\mathbf{F}_i^{(l)}||\mathbf{W}\mathbf{F}_j^{(l)}),
\end{equation}
where $\mathbf W\in \mathbb{R}^{d^{(l+1)}\times d^{(l)}}$ represents a learnable weight matrix, $\mathbf a^{(l)^T} \in \mathbb{R}^{2d^{(l+1)}}$ are the learnable attention coefficients and $||$ refers to the concatenation operation. 

A softmax function is then applied to the computed normalized attention scores such that,
\begin{equation}
    \alpha_{ij}^{(l)} = \frac{\text{exp}(e_{ij}^{(l)})}{\sum_{k\in \mathcal{N}{(i)}}\text{exp}(e_{ik}^{(l)})},
\end{equation}
with $\mathcal{N}{(i)}$ defining the neighborhood of the node $i$ and $\alpha_{ij}$ being the 
 obtained normalized attention score. 
 
We recall that the goal of the GCN subnet is to generate interdependent label classifiers. This means that each classifier must be predominated by the information related to the label it belongs to. Hence, the attention score of the node in question should be maximal. For this purpose, an additional step called self-importance mechanism is proposed for computing the attention-based adjacency matrix $\mathbf{B}^{(l)}=(b_{ij}^{(l)})_{i,j \in \mathcal{O}}$ as follows,

\begin{equation}
\label{eq:attention}
    \begin{cases}
       b_{ij}^{(l)}=\alpha_{ij}^{(l)} + \max \limits_{k \in \mathcal{O}}(\alpha_{ik}^{(l)})  \text{        if } i=j \\
       b_{ij}^{(l)} =\alpha_{ij}^{(l)} \text{     if }  i \neq j
    \end{cases}.
\end{equation}

\begin{figure*}[!t]
\begin{minipage}[b]{.33\linewidth}
  \centering
  \centerline{\includegraphics[width=5cm]{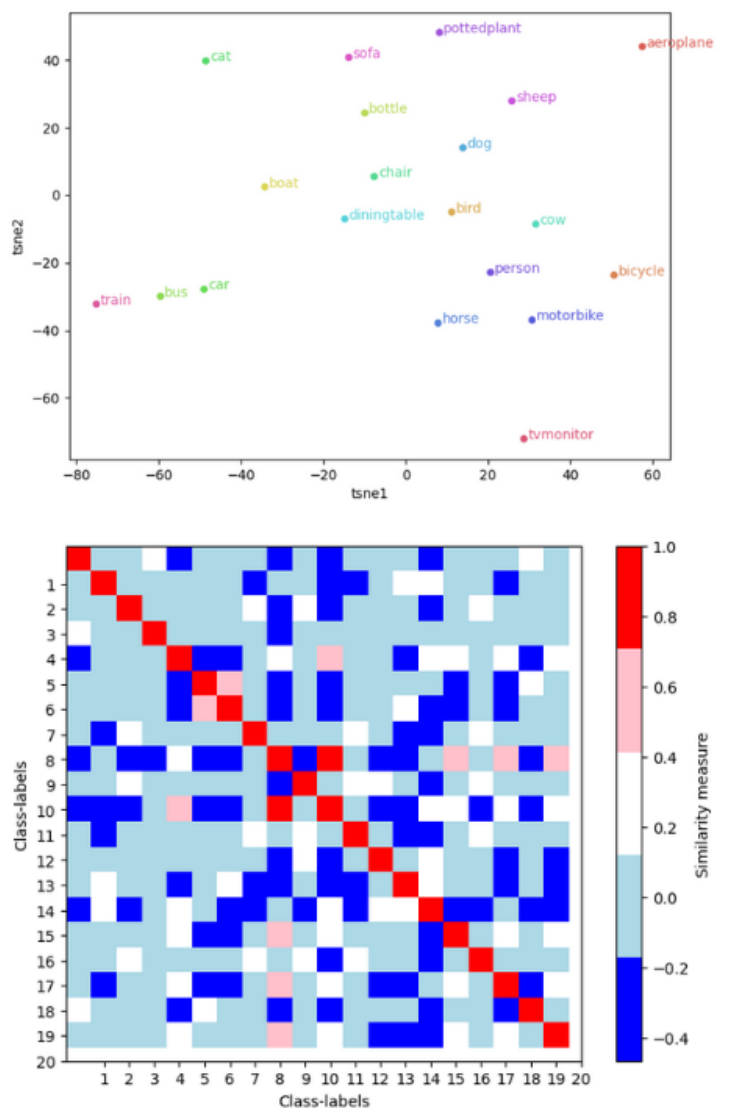}}
  \centerline{(a) Original}\medskip
\end{minipage}
\begin{minipage}[b]{.33\linewidth}
  \centering
  \centerline{\includegraphics[width=5cm]{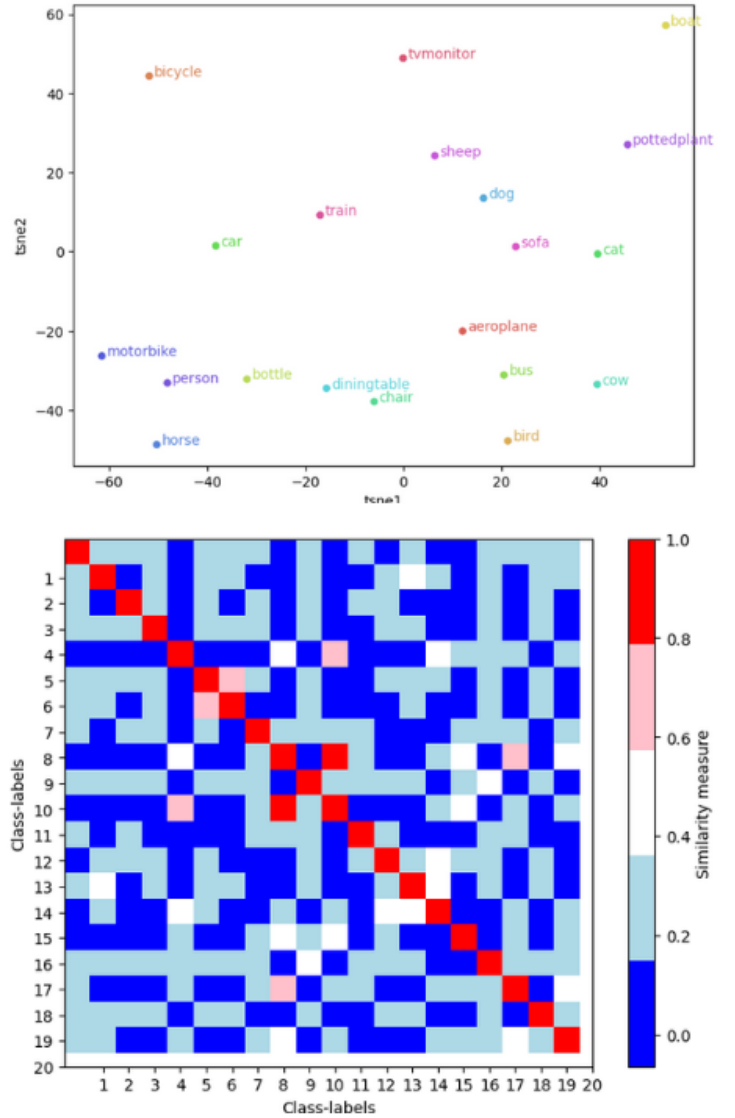}}
  \centerline{(b) ML-GCN}\medskip
\end{minipage}
\begin{minipage}[b]{.33\linewidth}
  \centering
  \centerline{\includegraphics[width=5cm]{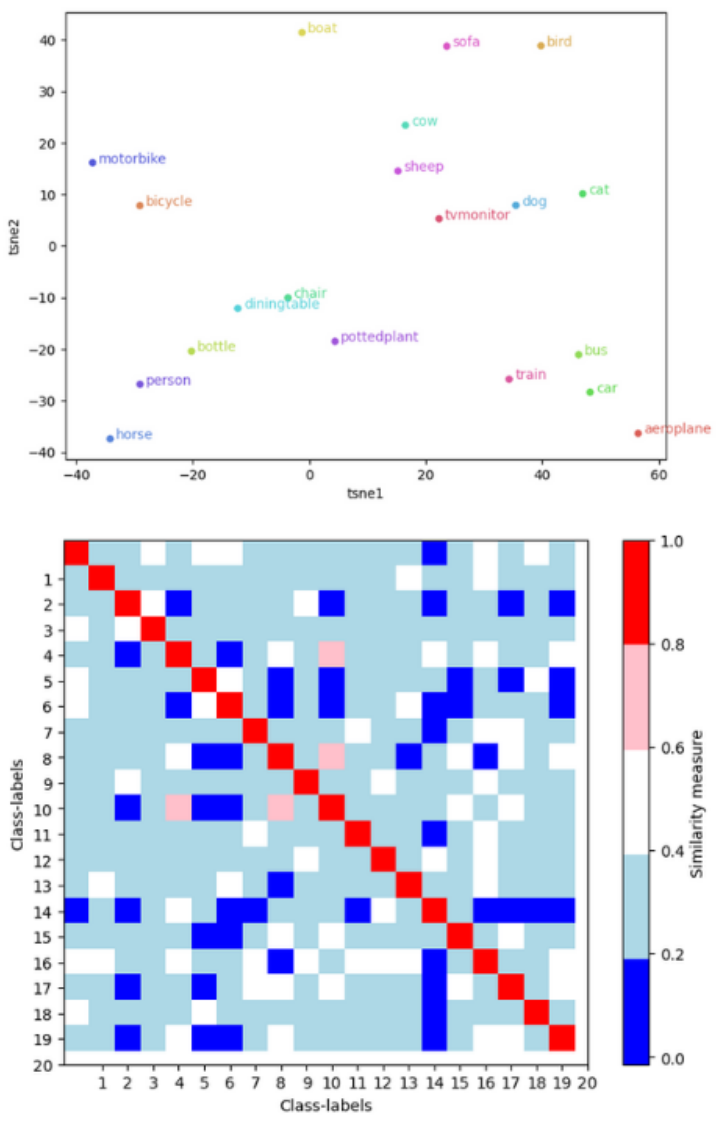}}
  \centerline{(c) Ours}\medskip
\end{minipage}
\hfill
\caption{Comparison of node feature similarity: The top row presents a tSNE visualization, while the bottom row illustrates cosine-similarity map between the graph nodes for VOC dataset: a) using the original image-based embeddings (before GCN), b) after applying two layers of standard GCN using the proposed architecture in ML-GCN~\citep{ml-gcn}, and c) after applying two layers of AGCN using our approach (i.e., ML-AGCN).}
\label{fig:node_sim}
\end{figure*}
 \subsubsection{Similarity-based adjacency matrix $(\mathbf{C})$}
 As illustrated in the top row of Fig.~\ref{fig:node_sim},  the GCN aggregation process tends to modify the node similarity in the original feature space~\citep{nodesim}. Subsequently, the second row illustrates a significant change in the similarity measure between the node features before and after the GCN operation (see Fig 4 a) and b), respectively). 
 {Although learned in an end-to-end manner, the attention-based graph $\mathbf B$ only quantifies the connectivity importance of each node pair and does not guarantee that the feature similarity is not lost through graph convolutions. Hence, we propose to incorporate an additional matrix $\mathbf{C}^{(l)}=(c_{ij}^{(l)})_{i,j \in \mathcal{O}}$ to preserve the node feature similarity.} It is obtained by calculating the cosine similarity $c_{ij}^{(l)}$ between each pair of vertices $(v_i$,$v_j)$ as follows,
 \begin{equation}
 \label{eq:cos}
     c_{ij}^{(l)} = \frac{\mathbf{F}_i^{(l)}.\mathbf{F}_j^{(l)}}{\|\mathbf{F}_i^{(l)}\|\|\mathbf{F}_j^{(l)}\|},
 \end{equation}
 where $\|.\|$ denotes the $L_2$ Euclidean norm. 
 
 \noindent
Finally, the output of the final layer $L$ denoted by $\mathbf F^L$ is used in Eq.~\eqref{eq_pred} for predicting the labels. As shown in Fig.~\ref{fig:node_sim} c), by using this strategy, the node similarity is relatively preserved, as compared to a standard GCN.

\section{Unsupervised Domain Adaptation for Multi-label Image Classification using ML-AGCN }
\label{sec:method_2}

\begin{figure*}[!t]
  \centering
  \centerline{\includegraphics[width=\linewidth]{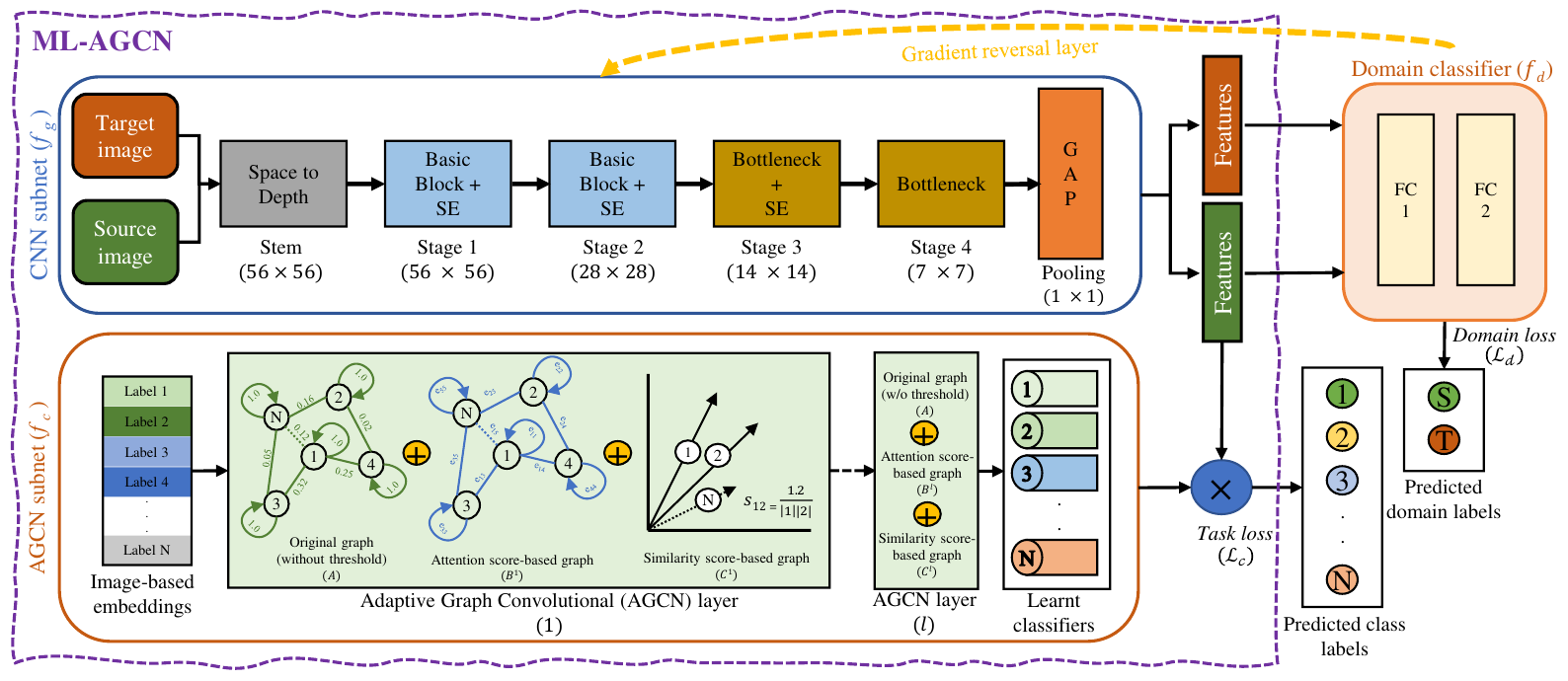}}
\caption{Architecture of the proposed DA-AGCN for multi-label image classification (best viewed in color).
Images from both source and target datasets are given as input to the CNN subnet that generates image features. The AGCN-subnet, similar to ML-AGCN~\citep{ml-agcn}, learns in an end-to-end manner the attention and similarity-based adjacency matrices $\mathbf B^{(l)}$ and $\mathbf C^{(l)}$, respectively, and generates accordingly inter-dependent label classifiers using only labeled source images. In addition, a domain classifier is considered.
}
\label{fig:arch_da}
\end{figure*}

The proposed architecture for multi-label image classification called DA-AGCN is illustrated in Fig.~\ref{fig:arch_da}. Similar to ML-AGCN, we make use of TResNet-based backbone $f_{g}$ to extract discriminative image features and an Adaptive Graph Convolutional Network $f_{c}$ to learn $N$ interdependent classifiers. Given source and target input images from $\mathcal{D}_s$ and $\mathcal{D}_t$, respectively, the goal of $f_{g}$ is to generate $D$-dim domain-invariant image features. The latter is used to predict the image labels, as depicted in Eq.~\eqref{eq_pred}. The label classification loss $\mathcal{L}_{c}$ is therefore defined as in Eq.~\eqref{eq:loss_cls}. 

For obtaining domain-invariant representations, a domain classifier $f_{d} \colon \mathbb {R}^{d_f} \to [\![ 0,1 ]\!]$ is employed and trained in an adversarial manner. 
Given the features $\mathbf X=f_{g}(\mathbf I)$ extracted from a given image $\mathbf I$, $f_{d}$  predicts the domain of the input image as follows, 

\begin{equation}
\begin{split}
       f_{d} \colon & \mathbb{R}^{d_f} \to [0,1]\\
  & \mathbf X \mapsto \hat{d}.
\end{split}
\end{equation}
\noindent 
where $\hat{d}$ is the predicted domain label of $\mathbf I$. Note that the ground-truth domain label $d=0$ if $\mathbf I$ is from the source domain and $d=1$ if sampled from the target domain. 

As in~\citep{dann}, a domain loss is defined as below,

\begin{multline}
\label{eq:domain_loss}
\mathcal{L}_{d}= \mathbb{E}_{f_{g}(\mathbf I_s) \sim \mathcal D_s}
 \log \frac{1}{f_d(f_g(\mathbf I_s))} + \\ \mathbb{E}_{f_{g}(\mathbf I_t) \sim \mathcal D_t}  \log \frac{1}{(1-f_d(f_g(\mathbf I_t))}.
 \end{multline}

\noindent
 Given $\mathcal{L}_{c}$ defined in Eq.~\eqref{eq:loss_cls}, the final objective function used to optimize the network is  $ E(\theta_{g}, \theta_{d}, \theta_{c})$ defined by,

\begin{equation}
\label{eq:lambda}
    E(\theta_{g}, \theta_{d}, \theta_{c}) = \mathcal{L}_{c}(\theta_{c},\theta_{g}) +\lambda \mathcal{L}_{d}( \theta_{d},\theta_{g}),
\end{equation}
where $\theta_{g}, \theta_{d}$ and $ \theta_{c}$ refer, respectively, to the parameters of $f_g$, $f_d$ and $f_c$ and $\lambda$ is a hyper-parameter defining the weight of $\mathcal L_d$.
The network is trained in an adversarial manner using the GRL for obtaining the optimal parameters $(\hat{\theta}_g,  \hat{\theta}_c, \hat{\theta}_d)$ such that,

\begin{equation}
\begin{split}
    (\hat{\theta}_g, \hat{\theta}_c) = \min_{\theta_{g},\theta_{c}} E(\theta_{g}, \theta_{c},\hat{\theta}_{d}),\\
    \hat{\theta}_{d} = \max_{\theta_{d}} E(\hat{\theta}_{g}, \hat{\theta}_c, \theta_{d})   .
\end{split}
\label{eq:minmax}
\end{equation}

\section{Experiments}
\label{sec:exp}
In this section, the experimental settings and results are presented and discussed for: (1) single-domain multi-label image classification (Section~\ref{exp-ml}); and (2) domain-adaptation for multi-label image classification (Section~\ref{exp-dom}).
\subsection{Multi-label Image Classification}
\label{exp-ml}
\subsubsection{Implementation details}
As discussed in Section~\ref{sec:method_1}, TResNet-M~\citep{tresnet} is used as the CNN backbone. In particular, the fully connected layer after the Global Average Pooling (GAP) layer is removed. The output of the GAP layer produces a 2048-dimensional latent image representation. The dimension of the AGCN output has also been set to 2048. For the node features, we use the image-based embeddings proposed in ~\citep{iml-gcn} instead of the usual word-based embeddings used in~\citep{ml-gcn}. The model is trained for 40 epochs using \textit{Adam}, with a maximum learning rate of $1e-4$ using a cosine decay.

\begin{table*}[!t]
\caption{Comparison with the state-of-the-art methods on the MS-COCO dataset. (Best and second-best performances are indicated in bold and are underlined, respectively).}
\centering
\resizebox{0.9\textwidth}{!}{%

\begin{tabular}{|l|l|c|c|c|cccccc|}
\hline
\label{table:coco-result}
\textbf{Category}                                      & \textbf{Method}     & \textbf{Resolution} & \textbf{\# params} & \textbf{mAP}                          & \textbf{CP}                  & \textbf{CR}                  & \textbf{CF1}                 & \textbf{OP}                  & \textbf{OR}                  & \textbf{OF1}                          \\
\hline
                                                       & Multi-Evidence~\citep{multi-evidence}      & 448x448             & 49.0               & -                                     & 80.4                         & 70.2                         & 74.9                         & 85.2                         & 72.5                         & 78.4                                  \\
                                                       & SRN~\citep{srn}                 & 224x224             & 48.0               & 77.1                                  & 81.6                         & 65.4                         & 71.2                         & 82.7                         & 69.9                         & 75.8                                  \\
                                                       & ResNet101~\citep{resnet}           & 448x448             & 44.5               & 77.3                                  & 80.2                         & 66.7                         & 72.8                         & 83.9                         & 70.8                         & 76.8                                  \\
                                                       & MCAR~\citep{mcar} & 448×448 & 44.9 & 83.8 & 85   & 72.1 & 78   & 88   & 73.9 & 80.3 \\
                                                        & MCAR~\citep{mcar} & 576x576 & 44.9 & 84.5 & 84.3 & 73.9 & 78.7 & 86.9 & 76.1 & 81.1 \\
\multirow{-6}{*}{CNN}                                  & TResNet-L~\citep{asl}           & 448x448             & 53.8               & 86.6                                  & \underline{{87.4}}                & 76.4                         & \underline{{81.4}}                         & \underline{{88.1}}                & 79.2                         & 81.8                                  \\
\hline
RNN                                                    & CNN-RNN~\citep{cnn-rnn}             & 224x224             & 66.2               & 61.2                                  & -                            & -                            & -                            & -                            & -                            & -                                     \\
\hline
                                                       & ML-GCN (1-layer)~\citep{ml-gcn}    & 448x448             & 43.1               & 80.9                                  & 82.9                         & 69.7                         & 75.8                         & 84.8                         & 73.6                         & 78.8                                  \\
                                                       & IML-GCN   (1-layer)~\citep{iml-gcn} & 448x448             & \textbf{29.5}               & 81.3                                  & 81.3                         & 72.2                         & 76.0                         & 86.7                         & 77.9                         & 82.1                                  \\
                                                       & ML-GCN~\citep{ml-gcn}              & 448x448             & 44.9               & 83.0                                  & 85.1                         & 72.0                         & 78.0                         & 85.8                         & 75.4                         & 80.3                                  \\
                                                       & {ML-GCN (TResNetM)}~\citep{ml-gcn} & {448×448} & {31.9} & {82.4} & {\underline{87.6}} & {66.6} & {75.2} & {\underline{91.0}} & {70.3} & {79.3}\\
                                                       & A-GCN~\citep{a-gcn}               & 512x512             & 44.1               & 83.1                                  & 84.7                         & 72.3                         & 78.0                         & 85.6                         & 75.5                         & 80.3                                  \\
                                                       & F-GCN~\citep{f-gcn}               & -                   & 44.3               & 83.2                                  & 85.4                         & 72.4                         & 78.3                         & 86.0                         & 75.7                         & 80.5                                  \\
                                                       & CFMIC~\citep{cfmic}               & 448×448             & 45.1               & 83.8                                  & 85.8                         & 72.7                         & 78.7                         & 86.3                         & 76.3                         & 81.0                                  \\
                                                       & SSGRL~\citep{ssgrl}               & 576x576             & 92.2               & 83.8                                  & \textbf{89.9}                & 68.5                         & 76.8                         & \textbf{91.3}                & 70.8                         & 79.7                                  \\
                                                       & FLNet~\citep{FLNet}               & -                   & 46.0               & 84.1                                  & 84.9                         & 73.9                         & 79.0                         & 85.5                         & 77.4                         & 81.1                                  \\
\multirow{-10}{*}{Graph-based}                          & IML-GCN~\citep{iml-gcn}             & 448x448             & 31.5               & 86.6                                  & 78.8                         & \textbf{82.6}                & 80.2                         & 79.0                         & \textbf{85.1}                & 81.9                                  \\
\hline
&{IDA~\citep{ida}} & 576x576 & 55.1 & 86.3 & - & - & 80.4 & - & - & 82.5 \\
                                   & MlTr-s~\citep{mltr}              & 384x384             & 33.0               & 83.9                                  & 82.8                         & 75.5                         & 77.3                         & 83                           & 78.5                         & 79.9                                  \\
                                   & STMG~\citep{stmg}                & 384x384             & 197.0              & 84.3                                  & 85.8                         & 72.7                         & 78.7                         & 86.7                         & 76.8                         & 81.5                                  \\
                                   & C-Tran~\citep{c-trans}              & 576x576             & 120.0              & 85.1                                  & 86.3                         & 74.3                         & 79.9                         & 87.7                         & 76.5                         & 81.7                                  \\
                                   & MlTr-m~\citep{mltr}              & 384x384             & 62.0               & 86.8                                  & 84                           & 80.1                         & \textbf{81.7}                & 84.6                         & 82.5                         & \underline{{83.5}}                                  \\
{\multirow{-6}{*}{Transformer-based}} & Ml-Decoder~\citep{ml-decoder}              & 448x448             & 51.3              & \textbf{88.1}                         & -                         & -                         & -                & -                         & -                         & -                         \\
\hline
\multicolumn{2}{|c|}{\textbf{Ours: ML-AGCN (1-layer)}}                         & 448×448             & \underline{{29.9}}               & 86.7                                  & 79.6                         & \underline{{82.4}}                & 80.7                         & 79.8                         & \underline{{84.5}}                & 82.1                                  \\
\multicolumn{2}{|c|}{\textbf{Ours: ML-AGCN (2-layers)}}                        & 448×448             & 35.9               & \underline{{86.9}} & 86.2 & 78.3 & \textbf{81.7} & 87.2 & 80.7 & \textbf{83.8}\\
\hline
\end{tabular}}
\end{table*}

\subsubsection{Datasets}  In the following, the datasets that have been used for the experiments are presented.
\paragraph{MS-COCO} MS-COCO~\citep{coco} is a widely used large-scale multi-label image dataset. It contains 80K training images and 40k testing images. Each image is annotated with multiple object labels from a total of 80 categories. 

\paragraph{VG-500} The VG-500 dataset~\citep{vg-500} is a well-known dataset for multi-label image classification. It includes 500 different objects as categories. The dataset comes with a training set of 98,249 images and a testing set of 10,000 images. 

\paragraph{PASCAL-VOC 2007} The PASCAL Visual Object Classes Challenge~\citep{voc} introduced in 2007 is one of the most commonly used multi-label image classification datasets. It contains about 10K image samples with 5011 and 4952 images as training and testing sets, respectively. The images show 20 different object categories with an average of 2.5 categories per image.

\subsubsection{Quantitative analysis}
\label{sec:exp_quant_agcn}
\paragraph{Comparison with state-of-the-art methods in terms of mAP and model size} We compare the performance of ML-AGCN with current state-of-the-art methods by reporting the mean Average Precision (mAP) as well as the number of model parameters. Additionally, similar to~\citep{iml-gcn, ml-gcn}, we report the following evaluation metrics on the MS-COCO dataset, namely, average per-Class Precision (CP), average per-Class Recall (CR), average per-Class F1-score (CF1),  average Overall Precision (OP), average overall recall (OR) and average Overall F1-score (OF1). 

Table~\ref{table:coco-result}, \ref{table:vg-results} and \ref{table:voc-results} report the quantitative comparison of the proposed approach with respect to state-of-the-art methods on the MS-COCO, the VG-500, and the VOC datasets, respectively. It can be clearly seen that our method achieves competitive results as compared to existing methods in terms of mAP while considerably reducing the model size. More specifically, similar to SSGRL, we achieve the best mAP performance on VOC-2007  while reducing the number of parameters from 92.2 to 35.8 million. 
 Moreover, ML-AGCN reaches the second-best mAP performance on MS-COCO and VG-500. 
However, as indicated in Table~\ref{table:coco-result} and Table~\ref{table:vg-results}, our method achieves the second-best performance on MS-COCO and VG-500.  Specifically, ML-decoder~\citep{ml-decoder} and C-Tran~\citep{c-trans} outperform our approach by 1.2\% and 0.5\% in mAP on MS-COCO and VG-500. This slight difference in performance might be explained by the fact that ML-decoder and C-Tran incorporate a noticeably higher number of parameters, i.e., 51.3 and 120 million, respectively, against only 35.9 and 32.1 million for ML-AGCN. Consequently, our approach achieves comparable performance while necessitating around 30\% and 70\% less parameters than ML-decoder and C-Tran, respectively. Such results show that our approach can maintain competitive performance while considerably reducing the network as compared to the state-of-the-art.  

It is also worth mentioning that ML-AGCN outperforms the ML-GCN~\citep{ml-gcn} baseline by 3.9\%, 6.3\%, and 0.5\% in terms of mAP on the MS-COCO, VG and VOC datasets, respectively, while keeping a comparable number of parameters. 

Moreover, for a fair comparison with this baseline, we re-train ML-GCN by replacing the original CNN backbone (ResNet101) with the same architecture used in our experiments, namely, TResNetM. The results that are reported in Table~\ref{table:coco-result}, Table~\ref{table:vg-results}, and Table~\ref{table:voc-results} show that ML-AGCN outperforms ML-GCN with a higher margin when using the same backbone, thereby confirming the relevance of the proposed adaptive graph convolution module.

In addition, we also report that the obtained performance when including only one layer in the graph-subnet of ML-GCN~\citep{ml-gcn}, IML-GCN~\citep{iml-gcn} and ML-AGCN. The obtained results confirm the relevance of the proposed adaptive learning. In fact, it can be seen that our method outperforms existing graph-based methods under this setting. More specifically, ML-AGCN achieves an improvement of  5.7\% and 5.3\% in terms of mAP when compared to ML-GCN and IML-GCN, respectively, on the MS-COCO dataset. Similarly, on the VG-500 benchmark, a significant increase of 20.7\% is recorded in terms of mAP in comparison to IML-GCN.

\begin{table}[t]
\caption{Comparison with the state-of-the-art methods on the VG-500 dataset.(Best and second to the best performance is indicated in bold and underline respectively).}
\label{table:vg-results}
\centering
\resizebox{0.48\textwidth}{!}{%
\begin{tabular}{|l|c|c|cccccc|}
\hline
\textbf{Method}                   & \textbf{\# params}     & \textbf{mAP}           & \textbf{CP}            & \textbf{CR}            & \textbf{CF1}           & \textbf{OP}            & \textbf{OR}            & \textbf{OF1}           \\
\hline
{ML-GCN (TResNetM)}~\citep{ml-gcn} & {33.2} & {26.6} & {30.7} & {7.7} & {10.3} & {\textbf{75.4}} & {10.0} & {17.7}\\
ResNet101~\citep{resnet}                & 44.5          & 30.9          & 39.1         & 25.6          & 31.0          & 61.4          & 35.9          & 45.4          \\
ML-GCN~\citep{ml-gcn}                   & 44.9          & 32.6          & 42.8          & 20.2          & 27.5          & 66.9 & 31.5          & 42.8          \\
TResNet-M~\citep{asl}                & \textbf{29.5} & 33.6          & -              & -               & -              & -              & -              & -              \\
IML-GCN~\citep{iml-gcn}                  & \underline{{32.1}} & 34.5          & -               & -              & -              & -              & -              & -              \\
SSGRL~\citep{ssgrl}                & 92.2          & 36.6          & -              & -              & -              & -              & -              & -              \\
    KGGR~\citep{vg-500}                 & 45.0          & 37.4          & \underline{{47.4}} & 24.7          & 32.5          & 66.9 & 36.5          & 47.2          \\
C-Tran~\citep{c-trans}                   & 120.0         & \textbf{38.4} & \textbf{49.8} & \underline{{27.2}} & \textbf{35.2} & 66.9 & \underline{{39.2}} & \underline{{49.5}} \\
\hline
\textbf{Ours: ML-AGCN (1-layer)}  & \underline{{32.1}}          & \underline{{37.9}} & 47.2          & \textbf{31.8} & \underline{{34.7}} & 64.1          & \textbf{42.1} & \textbf{50.8} \\
\textbf{Ours: ML-AGCN (2-layers)} & 37.4          & 37.1 & 47.3 & 24.7 & 29.0 & \underline{67.6} & 38.1 & 48.7\\
\hline
\end{tabular}}
\end{table}

\begin{table*}[!t]
\caption{Comparison with the state-of-the-art methods on the VOC-2007 dataset. (Best and second to the best performance is indicated in bold and underline respectively).}
\label{table:voc-results}
\resizebox{\textwidth}{!}{%
\begin{tabular}{|l|c|c|cccccccccccccccccccc|}
\hline
\textbf{Methods}             & \textbf{\# params} & \textbf{mAP}           & \textbf{aero}                & \textbf{bike}                & \textbf{bird}                & \textbf{boat}                & \textbf{bottle}              & \textbf{bus}                 & \textbf{car}                 & \textbf{cat}                 & \textbf{chair}               & \textbf{cow}           & \textbf{table}               & \textbf{dog}                 & \textbf{horse}               & \textbf{mbike}               & \textbf{person}              & \textbf{plant}               & \textbf{sheep}               & \textbf{sofa}                & \textbf{train}               & \textbf{tv}                  \\
\hline
CNN-RNN~\citep{cnn-rnn}   & 66.2      & 84.0          & 96.7                & 83.1                & 94.2                & 92.8                & 61.2                & 82.1                & 89.1                & 94.2                & 64.2                & 83.6          & 70                  & 92.4                & 91.7                & 84.2                & 93.7                & 59.8                & 93.2                & 75.3                & \textbf{99.7}       & 78.6                \\
VeryDeep~\citep{vgg}   & 138           & 89.7          & 98.9                & 95.0                & 96.8                & 95.4                & 69.7                & 90.4                & 93.5                & 96.0                & 74.2                & 86.6          & 87.8                & 96.0                & 96.3                & 93.1                & 97.2                & 70.0                & 92.1                & 80.3                & 98.1                & 87.0                \\
ResNet101~\citep{resnet}               & 44.5      & 89.9      & 99.5      & 97.7      & 97.8      & 96.4      & 65.7      & 91.8      & 96.1      & 97.6      & 74.2      & 80.9      & 85        & 98.4      & 96.5      & 95.9      & 98.4      & 70.1      & 88.3      & 80.2      & 98.9      & 89.2 \\
HCP~\citep{hcp}        & 138.0           & 90.9          & 98.6                & 97.1                & 98.0                & 95.6                & 75.3                & 94.7                & 95.8                & 97.3                & 73.1                & 90.2          & 80.0                & 97.3                & 96.1                & 94.9                & 96.3                & 78.3                & 94.7                & 76.2                & 97.9                & 91.5                \\
ML-GCN~\citep{ml-gcn}            & {44.9} & {94.0} & {99.5} & {\underline{98.5}} & {98.6} & {98.1} & {80.8} & {94.6} & {97.2} & {98.2} & {82.3} & {95.7} & {86.4} & {98.2} & {98.4} & {96.7} & {\underline{99.0}} & {84.7} & {96.7} & {84.3} & {98.9} & {93.7} \\
{ML-GCN (TResNetM)}~\citep{ml-gcn} & 31.8 & 94.1 & \underline{99.8} & 98.0 & \textbf{98.9} & 98.0 & 80.4 & 95.9 & 96.0 & 97.8 & 83.4 & \underline{98.6} & 86.4 & 98.6 & \textbf{99.0} & 95.1 & 98.8 & 82.7 & 98.8 & 84.5 & \textbf{99.7} & 92.0 \\
F-GCN~\citep{f-gcn}               & -          & 94.1          & 99.5                & \underline{{98.5}} & 98.7                & \underline{{98.2}} & 80.9                & 94.8                & 97.3                & 98.3                & 82.5                & 95.7          & 86.6                & 98.2                & 98.4                & 96.7                & \underline{{99.0}} & 84.8                & 96.7                & 84.4                & 99.0                & 93.7                \\
FLNet~\citep{FLNet}               & 46.0           & 94.4          & 99.6                & 98.1                & \textbf{98.9}       & 97.9                & 84.6                & 95.3                & 96.2                & 96.5                & \textbf{85.6}       & 96.1          & 87.2                & 97.7                & 98.6                & \underline{{97.0}} & 98.1                & \underline{{86.5}} & 97.4                & 86.5                & 98.8                & 90.8                \\
TResNet-L~\citep{asl}           & 53.8      & 94.6          & -                 & -                 & -                 & -                 & -                 & -                 & -                 & -                 & -                 & -           & -                 & -                 & -                 & -                 & -                 & -                 & -                 & -                 & -                 & -                 \\
CFMIC~\citep{cfmic}               & 45.1      & 94.7          & 99.7 & \underline{{98.5}} & \underline{{98.8}} & \textbf{98.3}       & 83.9                & 96.5                & \underline{{97.5}} & \underline{{98.8}} & 83.1                & 96.1          & 87.4                & 98.6                & \underline{{98.9}} & \textbf{97.2}       & \underline{{99.0}} & 85.4                & 97.1                & 84.9                & \underline{{99.2}} & 94.2                \\
MCAR~\citep{mcar}      & 44.9           & \underline{{94.8}}    & 99.7 & \textbf{99.0}       & 98.5                & \underline{{98.2}} & \underline{{85.4}} & \textbf{96.9}       & 97.4                & \textbf{98.9}       & \underline{{83.7}} & 95.5          & \underline{{88.8}} & \underline{{99.1}} & 98.2                & 95.1                & \textbf{99.1}       & 84.8                & 97.1                & \underline{{87.8}} & 98.3                & \textbf{94.8}       \\
SSGRL~\citep{ssgrl} & 92.2      & \textbf{95.0} & 99.7 & 98.4                & 98.0                & 97.6                & \textbf{85.7}       & 96.2                & \textbf{98.2}       & \textbf{98.8}       & 82.0                & 98.1 & \textbf{89.7}       & 98.8                & 98.7                & \underline{{97.0}} & \underline{{99.0}} & \textbf{86.9}       & \underline{{98.1}} & 85.8                & 99.0                & 93.7                \\
\hline
\textbf{Ours ML-AGCN (1-layer)}   & 29.5          & 94.5          &                     &                     &                     &                     &                     &                     &                     &                     &                     &               &                     &                     &                     &                     &                     &                     &                     &                     &                     &                    \\
\textbf{Ours ML-AGCN (2-layers)}  & 35.8          & \textbf{95.0} & \textbf{99.9}       & 98.0                & 98.5                & 98.0                & 81.6                & \underline{{96.8}} & 96.6                & 98.2                & \textbf{85.6}       & \textbf{99.4} & 88.2                & \textbf{99.2}       & \textbf{99.0}       & 96.5                & 98.8                & 84.8                & \textbf{99.5}       & \textbf{88.1}       & 98.9                & \underline{{94.5}} \\

\hline
\end{tabular}}
\end{table*}

\paragraph{Ablation study}

In order to analyze the impact of each adjacency matrix used in our approach, namely, the attention-based matrix $\mathbf B$ and similarity-based matrix $\mathbf C$, an ablation study is carried out as shown in Table~\ref{table:ablation_mlic}. It can be noted that by considering $\mathbf B$ in addition to $\mathbf A$ (without threshold), an improvement of 5.1\%, 20.5\%, and 0.04\% in terms of mAP is made, respectively, on MS-COCO, VG-500, and VOC-2007. The magnitude of improvement seems dependent on the number of classes contained in the considered dataset. In fact, while VG-500 is formed by 500 classes, MS-COCO and VOC-2007 are respectively composed of 80 and 20 categories. This highlights the importance of adaptively modeling label correlations, especially when dealing with a large number of classes, which is likely to occur in a practical scenario. An additional mAP improvement of 0.1\%, 0.4\%, and 0.17\% can be seen when including $\mathbf C$ in the AGCN subnet. In conclusion, it can be noted that $\mathbf B$ contributes more importantly to the resulting enhancement. This could be explained by the fact that $\mathbf C$ is less needed, as only two layers are considered in the graph subnet. As shown in Fig 4., while the use of C allows to better preserve the feature similarity as, the latter is not completely lost through layers when using 2 standard GCN convolutions.

\begin{table}[!t]
\caption{The ablation of adaptively learning $\mathbf{B}$ and $\mathbf{C}$ for multi-label image classification in single domain.}
\label{table:ablation_mlic}
\resizebox{0.48\textwidth}{!}{%
\begin{tabular}{|l|c|c|c|}
\hline
\multirow{2}{*}{\textbf{Method}} & \multicolumn{3}{c|}{\textbf{mean Average Precision (mAP \%)}}\\
\cline{2-4}

& \textbf{MS-COCO} & \textbf{VG-500} & \textbf{VOC-2007} \\
\hline
ML-AGCN (A) & 81.1  & 17 & 94.27  \\
ML-AGCN (A+B)  & 86.6 \textbf{(+5.1\%)}  & 37.5 \textbf{(+20.5\%) }  & 94.31 \textbf{(+0.04\%) }\\
ML-AGCN (A+B+C)  & 86.7\textbf{ \textbf{(0.1\%)} } & 37.9 \textbf{(+0.4\%)}   & 95.0\textbf{ (+0.69\%) } \\
\hline
\end{tabular}}
\end{table}

\begin{table}[!t]
\centering
\caption{{Hyper-parameter analysis: comparison of the performance of ML-AGCN when varying the number of AGCN layers. Best performance is indicated in \textbf{bold}.}}
\label{table:agcn_layers}
\resizebox{0.48\textwidth}{!}{%
\begin{tabular}{|l|c|c|c|c|}
\hline
\multicolumn{1}{|c|}{\multirow{2}{*}{\textbf{\#   layers}}} & \multirow{2}{*}{\textbf{\# params (millions)}} & \multicolumn{3}{c|}{\textbf{mean Average Precision   (\%)}} \\
\cline{3-5}
\multicolumn{1}{|c|}{}                             &                            & MS-COCO           & VG-500          & VOC-2007            \\
\hline
1-layer                                          & 29.9                       & 86.7           & 37.9            & 94.5           \\
2-layers                                         & 35.9                       & \textbf{86.9}           & \textbf{37.1}            & \textbf{95.0}           \\
3-layers                                         & 37.3                       & 85.7           & 24.6            & 94.7           \\
4-layers                                         & 40.7                       & 85.4           & -              & 94.6           \\
\hline
\end{tabular}}
\end{table}

\paragraph{Hyper-parameter analysis}
Table~\ref{table:agcn_layers} reports the mean Average Precision (mAP) for MS-COCO, VG-500, and VOC-2007 datasets, when varying the total number of layers in ML-AGCN. The best results are generally obtained when only two layers are considered. However, it should be noted that the results are very stable for MS-COCO and VOC-2007, with a variation lower than 1.5\% for all the configurations.

\subsubsection{Qualitative analysis}
\begin{figure*}[!t]
\centering
\includegraphics[width=\linewidth]{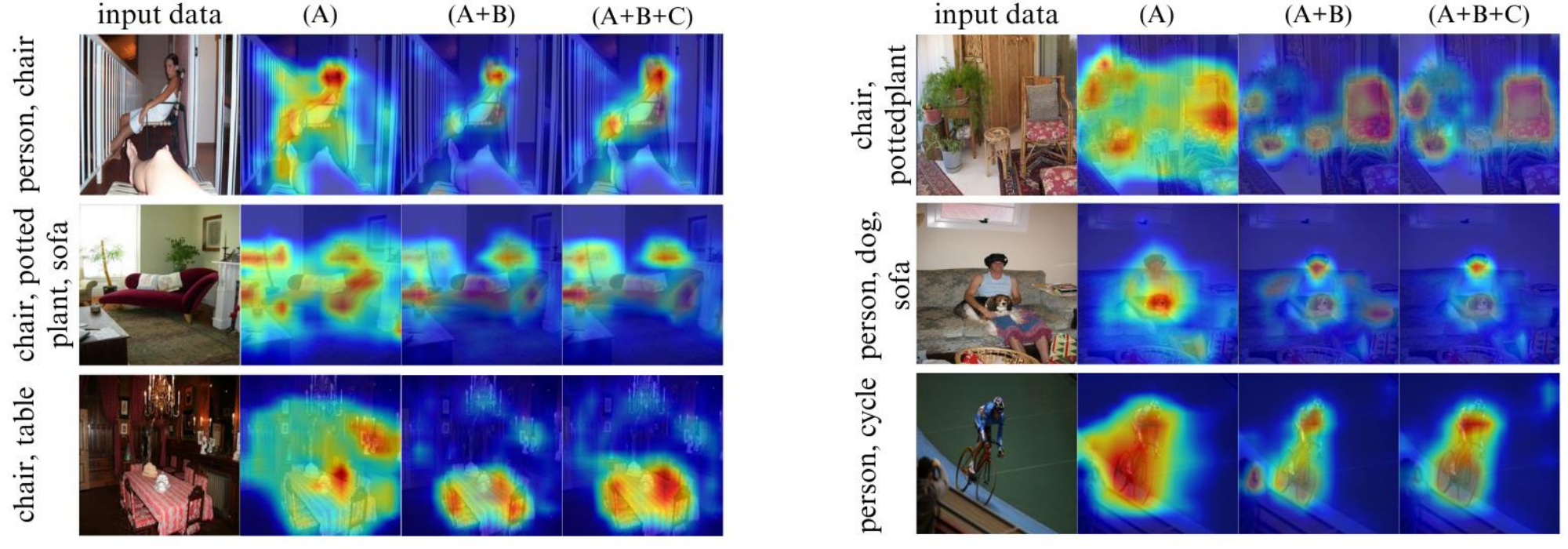}
\caption{Grad-CAM visualization of the predictions with ML-AGCN {~\citep{ml-agcn}} using samples from the VOC dataset using only $\mathbf A$, then   $\mathbf{A}$ and $\mathbf{B}$, and finally $\mathbf{A}$, $\mathbf{B}$ and  $\mathbf{C}$.}
\label{fig:qual_voc}
\end{figure*}

In Fig.~\ref{fig:qual_voc}, the Gradient-weighted Class Activation Mapping (Grad-CAM)~\citep{grad-cam} is visualized for some examples from the VOC dataset. While the first column of images represents the input images, the second, third, and fourth columns show, respectively, the Grad-cam visualization from a model trained using only  $\mathbf{A}$, $\mathbf{A}$ and $\mathbf{B}$, and finally $\mathbf{A}$, $\mathbf{B}$ and $\mathbf{C}$. These qualitative results are in line with the quantitative ones. As shown in Fig.~\ref{fig:qual_voc}, the use of $\mathbf B$ allows activating more precisely the regions of interest in the image, while the contribution of $\mathbf{C}$ in this refinement is less impressive but remains visible. 

\subsection{Unsupervised Domain Adaptation for Multi-label Image Classification}
\label{exp-dom}
\subsubsection{Implementation details}
We reproduce the results of current state-of-the-art methods due to the limited availability of DA approaches for multi-label image classification. In particular, we first consider standard multi-label image classification methods (without DA) and refer to them as \textit{MLIC}. Since no target images were employed during the learning process, this is equivalent to source-only training. {Second, we evaluate the zero-shot performance of existing Vision Language Models (VLM) on the target dataset referred to as VLM in our experiments. For this purpose, we use a pre-trained Contrastive Language-Image Pre-Training (CLIP~\citep{clip} model with a ViT-B/32 transformer-based backbone. As commonly done in MLIC~\citep{clip4mlic}, we compute a similarity score, given an input image to VLM, between the predicted text features and ground truth features of the considered labels. The application of a sigmoid activation function to these scores results in the presence probabilities of the considered objects.}

Lastly, we reproduce the outcomes of two state-of-the-art domain adaptation methods, namely DANN~\citep{dann} and DA-MAIC~\citep{da-maic}, that we refer as \textit{DA} in our experiments. Moreover, we replace the multi-label softmargin-loss with the traditional cross-entropy loss when training DANN. We generate Glove-based word embeddings~\citep{glove} as node features for training DA-MAIC. Additionally, in order to showcase the effectiveness of the proposed AGCN subnet, we present the results of DANN and DA-MAIC by considering the same backbone as ours: an additional experiment based on TResNet-M instead of the traditional ResNet101 is carried out. The images have been resized to 224 $\times$ 224, unless stated differently. Our domain classifier includes one hidden layer of dimension 1024. A maximum learning rate of $1e-4$ using a cosine decay is considered. The model is trained for a total of 40 epochs or until convergence.

\subsubsection{Datasets}
Similar to DA-MAIC~\citep{da-maic}, we use three multi-label aerial image datasets in our experiments, namely, AID~\citep{aid-ml}, UCM~\citep{ucm-ml} and DFC15~\citep{dfc}. Additionally, due to the limited number of suitable datasets for the task of DA in MLIC, we convert two well-known object detection datasets, initially used for DA in the context of object detection, to multi-label annotations, namely PASCAL VOC 2007~\citep{voc} and Clipart1k~\citep{clipart}.

\paragraph{AID multi-label aerial dataset} The original AID dataset~\citep{aid}  contains 10000 high-resolution aerial images. The images cover a total of 30 categories. A multi-label version of this dataset was produced in~\citep{aid-ml}, where 3000 aerial images from the original AID dataset have been selected and assigned with multiple object labels. In total, they include 17 labels: airplane, sand, pavement, buildings, cars, chaparral, court, trees, dock, tank, water, grass, mobile-home, ship, bare-soil, sea, and field. 80\% and 20\%  of the images have been used respectively for training and testing~\citep{aid-ml}.

\paragraph{UCM multi-label aerial dataset} UCM multi-label dataset~\citep{ucm-ml} is derived from the UCM dataset~\citep{ucm}. It consists of images showing 21 land-use classes. The images have a resolution of 256 x 256 pixels. Later in~\citep{ucm-ml}, 2100 of these aerial images were annotated with multiple tags in order to generate a multi-label aerial image dataset. This dataset shares the same number of labels as AID multi-label dataset~\citep{aid-ml} i.e., 17 labels. In our experiments, 80\% and 20\% of data are respectively used for training and testing. 

\paragraph{DFC15 multi-label aerial dataset} The DFC15 multi-label dataset~\citep{dfc} was initially introduced in 2015. It has a total of 3342 high-resolution image samples and includes 8 object labels. In our experiments, the 6 categories in common with UCM and AID datasets, including water, grass, building, tree, ship, and car, are considered. 80\% and 20\% are respectively used for training and testing.

\paragraph{VOC and Clipart1k datasets} The Clipart1k~\citep{clipart} dataset contains 20 object categories, similar to PASCAL-VOC 2007~\citep{voc}. We create a multi-label annotation for each image by considering the category of each object bounding box. Clipart1k provides a total of 1000 image samples. 50\% and 50\% are respectively used for training and testing.

\begin{table*}[!t]
\caption{Comparison with the state-of-the-art in terms of mAP and number of model parameters using two settings, i.e., AID $\rightarrow$ UCM and UCM $\rightarrow$ AID. (Best performance is indicated in bold and second-best performance is underlined).}
\label{table:aid_ucm_results_both}
\resizebox{\textwidth}{!}{%
\begin{tabular}{|c|l|l|c|c|cccccc|c|cccccc|}
\hline
\multirow{2}{*}{\textbf{Category}} & \multicolumn{1}{c|}{\multirow{2}{*}{\textbf{Backbone}}} & \multirow{2}{*}{\textbf{Method}} & \multirow{2}{*}{\textbf{\# params}} & \multicolumn{7}{c|}{\textbf{AID $\rightarrow$ UCM}}                                                                                                   & \multicolumn{7}{c|}{\textbf{UCM $\rightarrow$ AID}}                                                                                                   \\
\cline{5-18}
                                   & \multicolumn{1}{c|}{}                                   &                                  &                                     & \textbf{mAP}        & \textbf{CP}            & \textbf{CR}            & \textbf{CF1}           & \textbf{OP}            & \textbf{OR}            & \textbf{OF1}         & \textbf{mAP}        & \textbf{CP}            & \textbf{CR}            & \textbf{CF1}           & \textbf{OP}            & \textbf{OR}            & \textbf{OF1}       \\
                                   \hline
\multirow{6}{*}{MLIC}              & ResNet50                                               & RESNET~\citep{resnet}                           & 23.5                                & 54.5                & 57.9                & 42.3                & 43.2                & \underline{{70.2}} & 67.1                & 68.6                & 50.3                & 57.2 & 26.8                & 32.2                & \underline{{90.0}} & 44.2                & 59.3                \\
 \cline{2-18}
                                   & \multirow{2}{*}{ResNet101}                                              & RESNET~\citep{resnet}                           & 42.5                                & 57.5 & \textbf{60.0}       & 47.5                & 47.0                & 69.1                & 71.5                & \textbf{70.3}       & 51.7                & 50.6                & 29.6                & 33.9                & 88.0                & 48.5                & 62.5                \\
                                   &                                               & ML-GCN~\citep{ml-gcn}                           & 44.9                                & 53.7                & 55.3                & 44.3                & 45.9                & 70.2                & 68.7                & \underline{{69.4}} & 51.3                & 50.1                & 29.9                & 34.0                & 88.0                & 49.7                & 63.6                \\
                                   \cline{2-18}
                                   & \multirow{3}{*}{TResNet-M} & {ML-GCN}~\citep{ml-gcn} &{31.8} & {53.5} & {57.8} & {40.5} & {41.1} & {64.2} & {70.4} & {{{67.2} }} & {52.5} & {51.1} & {25.8} & {31.9} & {\textbf{91.6}} & {40.8} & {56.5}\\
                                   &                                              & ASL~\citep{asl}                              & 29.4                                & 55.4                & 48.7                & 52.8                & 47.1                & 58.7                & 79.1                & 67.4                & {54.1} & 54.5                & 40.2                & \underline{41.9}                & 85.4                & 65.1                & 73.9 \\
                                   &                                              & ML-AGCN~\citep{ml-agcn}                          & 36.6                                & 55.2                & 36.6                & {\textbf{64.9}}       & 45.1                & 45.0                & \textbf{88.1}       & 59.6                & 52.1                & 48.2                & \textbf{47.4}       & \textbf{42.9} & 77.1                & \textbf{79.8}       & \textbf{78.4}       \\
                                   \hline
{VLM} &{ViT-B/32}  & {CLIP~\citep{clip}} & {151.3} & {42.2} & {45.8} & {25.2} & {28.2} & {51.2} & {23.6} & {32.3} & {42.4} & {44.8} & {21.8} & {22.4} & {54.1} & {15.9} & {24.5}\\
\hline
\multirow{6}{*}{DA}                & \multirow{2}{*}{ResNet101}                                              & DANN~\citep{dann}                             & 42.5                                & 55.3                & 54.7                & 57.9                & \underline{52.8}       & 59.0                & 81.6                & 68.5                & 50.5                & \underline{60.4}       & 24.9                & 32.3                & 89.2                & 42.2                & 57.3                \\
                                   &                & DA-MAIC~\citep{da-maic}                          & 44.9                                & 49.7                & 52.4                & 48.7                & 45.4                & 56.9                & 72.5                & 63.8                & 48.7                & 51.6                & \underline{40.9}                & 41.5                & 78.9                & 65.5 & 71.6                \\
                                   \cline{2-18}
                                   & \multirow{4}{*}{TResNet-M}                             & DANN~\citep{dann}                             & 29.4                                & 52.5                & \underline{{59.1}} & 31.6                & 36.3                & \textbf{70.9}       & 53.7                & 61.1                & 51.6                & 52.1                & 23.2                & 27.9                & 83.2                & 27.8                & 41.7                \\
                                   &                                                        & DA-MAIC~\citep{da-maic}                          & 31.8                                & 54.4                & 55.3                & 37.5                & 38.6                & 68.0                & 67.9                & 67.9                & 50.5                & 51.8                & 22.9                & 29.0                & \textbf{91.6}       & 35.2                & 50.8                \\
                                   & &{DDA-MLIC}~\citep{dda-mlic}    & {29.4}            & {\textbf{63.2}} & {52.5} & {\underline{63.7}} & {\textbf{55.1}} & {59.4} & {\underline{82.8}} & {69.2} & {\underline{54.9}} & {53.9} & {30.4} & {35.5} & {84.6} & {41.0} & {55.3}\\
                                   \cline{3-18}
                                   &                                                        & \textbf{DA-AGCN (Ours)}          & 36.6                                & \underline{59.0}       & 54.0                & {{59.4}} & {{52.3}} & 57.1                & {{82.3}} & 67.4                & {\textbf{57.2}} & {\textbf{61.2}} & {40.0} & {41.8} & {86.1} & {\underline{66.9}} & {\underline{75.3}}   \\
                                   \hline
\end{tabular}}
\end{table*}

\begin{table*}[!t]
\caption{Comparison with the state-of-the-art in terms of mAP and number of model parameters using two settings, i.e., UCM $\rightarrow$ DFC and AID $\rightarrow$ DFC. (Best performance is indicated in bold and second-best performance is underlined).}
\label{table:dfc_results_comb}
\resizebox{\textwidth}{!}{%
\begin{tabular}{|c|l|l|c|c|cccccc|c|cccccc|}
\hline
\multirow{2}{*}{\textbf{Category}} & \multicolumn{1}{c|}{\multirow{2}{*}{\textbf{Backbone}}} & \multirow{2}{*}{\textbf{Method}} & \multirow{2}{*}{\textbf{\# params}} & \multicolumn{7}{c|}{\textbf{AID   $\rightarrow$ DFC}}                                                                                                   & \multicolumn{7}{c|}{\textbf{UCM   $\rightarrow$ DFC}}                                                                                                   \\
\cline{5-18}
                                   & \multicolumn{1}{c|}{}                                   &                                  &                                     & \textbf{mAP}        & \textbf{CP}            & \textbf{CR}            & \textbf{CF1}           & \textbf{OP}            & \textbf{OR}            & \textbf{OF1}         & \textbf{mAP}        & \textbf{CP}            & \textbf{CR}            & \textbf{CF1}           & \textbf{OP}            & \textbf{OR}            & \textbf{OF1}          \\
                                   \hline
\multirow{6}{*}{MLIC}              & ResNet50                                               & RESNET~\citep{resnet}                           & 23.5                                & 52.9                & 52.3                & 47.1                & 44.2                & {52.3}       & 50.5                & 51.3                & 67.7                & 58.5                & 31.9                & 34.9                & 65.9                & 39.1                & 49.1                \\
\cline{2-18}
                                   & \multirow{2}{*}{ResNet101}                                              & RESNET~\citep{resnet}                           & 42.5                                & 56.9                & 52.9                & 61.5                & 48.7                & 46.1                & 63.7                & 53.5                & 66.4                & {{74.4}} & 31.2                & 36.9                & {{67.2}} & 37.2                & 47.9                \\
                                   &                                               & ML-GCN~\citep{ml-gcn}                           & 44.9                                & 58.9                & {{56.7}} & 57.9                & 45.8                & 45.7                & 65.0                & 53.7                & 64.6                & 72.4                & 32.0                & 35.6                & 64.4                & 38.9                & 48.5                \\
                                   \cline{2-18}
                                   & \multirow{3}{*}{TResNet-M} & {ML-GCN}~\citep{ml-gcn} &{31.8} & {53.5} & {{ 52.0}} & {47.9} & {41.6} & {\underline{52.7}} & {51.1} & {{ 51.9}} & {66.6} & {60.2} & {35.0} & {38.1} & {{64.9}} & {39.4} & {49.0} \\
                                   &                                              & ASL~\citep{asl}                              & 29.4                                & 56.1                & 49.6                & 68.4                & 49.9                & 43.5                & 74.1                & 54.8                & 68.9                & 66.3                & 53.1                & 44.0                & 52.6                & 57.0                & 54.7                \\
                                   &                                             & ML-AGCN~\citep{ml-agcn}                          & 36.6                                & 51.6                & 41.5                & \textbf{83.8}       & {{52.3}} & 40.2                & \textbf{88.7}       & 55.3                & {{70.3}} & 68.4                & \underline{{56.1}} & 47.8                & 53.8                & \underline{{58.5}} & 56.0                \\
                                   \hline
                                   {VLM} &{ViT-B/32}  & {CLIP~\citep{clip}} & {151.3} & {\underline{64.3}} & {\textbf{88.1}} & {22.3} & {31.4} & {\textbf{78.7}} & {22.0} & {34.4} & {63.8} & {\textbf{88.4}} & {22.7} & {31.3} & {\textbf{79.5}} & {22.5} & {35.1}\\
\hline
\multirow{5}{*}{DA}                & \multirow{2}{*}{ResNet101}                                              & DANN~\citep{dann}                             & 42.5                                & \underline{{64.3}} & \underline{57.0}       & 65.9                & 51.4                & 48.0 & 65.9                & {{55.6}} & 67.2                & 72.8                & 44.7                & 48.7                & 56.3                & 49.1                & 52.4                \\
                                   &                                               & DA-MAIC~\citep{da-maic}                          & 44.9                                & 50.1                & 54.5                & 45.8                & 38.9                & 44.2                & 48.8                & 46.4                & 65.6                & 69.7                & 54.2                & \underline{{52.8}} & 57.1                & 58.4                & \underline{{57.7}} \\
                                   \cline{2-18}
                                   & \multirow{3}{*}{TResNet-M}                             & DANN~\citep{dann}                             & 29.4                                & 43.0                & 40.7                & 13.6                & 19.3                & 46.0                & 15.6                & 23.3                & 64.1                & \underline{77.3}       & 22.6                & 30.1                & \underline{68.6}       & 26.5                & 38.2                \\
                                   &                                                        & DA-MAIC~\citep{da-maic}                          & 31.8                                & 55.4                & 49.8                & 60.4                & 44.7                & 47.3                & 64.1                & 54.4                & 65.8                & 71.4                & 39.3                & 39.7                & 59.9                & 44.6                & 51.1                \\
                                   & & {{DDA-MLIC}}~\citep{dda-mlic}                  & {29.4}            & {62.1}& {47.6}& {75.5}& {\underline{55.3}} & {48.9}& {76.2}& {\textbf{59.6}} & {\underline{70.6}}& {67.2}& {55.7}& {{49.3}} & {55.0}& {58.4}& {{56.6}} \\
                                   \cline{3-18}
                                   &                                                        & \textbf{DA-AGCN (Ours)}          & 36.6                                & \textbf{65.7}       & 51.8                & \underline{{78.1}} & \textbf{55.7}       & 45.2                & \underline{{80.8}} & \underline{58.0}       & \textbf{76.5}       & 68.5                & \textbf{61.7}       & \textbf{59.0}       & 60.0                & \textbf{60.2}       & \textbf{60.1}      \\
                                   \hline
\end{tabular}}
\end{table*}

\begin{table*}[!t]
\caption{Comparison with the state-of-the-art in terms of mAP and number of model parameters using the two settings, i.e., VOC $\rightarrow$ Clipart and Clipart $\rightarrow$ VOC. (Best performance is indicated in bold and second-best performance is underlined).}
\label{table:voc_results_comb}
\resizebox{\textwidth}{!}{%
\begin{tabular}{|c|l|l|c|c|cccccc|c|cccccc|}
\hline
\multirow{2}{*}{\textbf{Category}} & \multicolumn{1}{c|}{\multirow{2}{*}{\textbf{Backbone}}} & \multirow{2}{*}{\textbf{Method}} & \multirow{2}{*}{\textbf{\# params}} & \multicolumn{7}{c|}{\textbf{VOC $\rightarrow$ Clipart}}                                                     & \multicolumn{7}{c|}{\textbf{Clipart   $\rightarrow$ VOC}}                                                     \\
\cline{5-18}
                                   & \multicolumn{1}{c|}{}                                   &                                  &                                     & \textbf{mAP}  &\textbf{CP}            & \textbf{CR}            & \textbf{CF1}           & \textbf{OP}            & \textbf{OR}            & \textbf{OF1}         & \textbf{mAP}        & \textbf{CP}            & \textbf{CR}            & \textbf{CF1}           & \textbf{OP}            & \textbf{OR}            & \textbf{OF1}     \\
                                   \hline
\multirow{6}{*}{MLIC}              & ResNet50                                             & RESNET~\citep{resnet}                             & 23.5                                & 42.0          & 57.6          & 15.3          & 22.5          & 82.3          & 25.8          & 39.3          & 45.4          & 40.3          & 9.8           & 13.0          & 84.7 & 25.5          & 39.2          \\
\cline{2-18}
                                   & \multirow{2}{*}{ResNet101}                                              & RESNET~\citep{resnet}                             & 42.5                                & 38.0          & 64.8          & 14.3          & 22.5          & 82.3          & 18.3          & 29.9          & 50.1          & 66.2          & 17.5          & 25.5          & 83.9          & 29.6          & 43.7          \\
                                   &                                                 & ML-GCN~\citep{ml-gcn}                           & 44.9                                & 43.5          & 62.5          & 20.3          & 28.4          & 86.6          & 27.8          & 42.1          & 43.1          & 57.9          & 21.0          & 26.8          & 73.5          & 30.6          & 43.2          \\
                                   \cline{2-18}
                                   & \multirow{3}{*}{TResNet-M} & {ML-GCN}~\citep{ml-gcn}  &{31.8} & {53.0} & {{ 75.3}} & {26.9} & {37.3} & {{90.3}} & {31.3} & {{ 46.4}} & {67.6} & {{80.8}} & {39.9} & {50.8} & {\underline{88.8}} & {46.0} & {60.6} \\
                                   &                                             & ASL~\citep{asl}                              & 29.4                                & 56.8 & 72.0          & 38.5 & 47.6 & 82.8          & 45.7 & 58.9 & 64.2          & 69.0          & 30.7 & 37.3 & 80.0          & 45.7 & 58.2 \\
                                   &                                             & ML-AGCN~\citep{ml-agcn}                          & 36.6                                & 53.7          & 75.5          & 35.5          & 44.4          & 79.1          & 39.9          & 53.1          & 38.0          & 45.5          & 25.1          & 28.2          & 61.8          & 36.6          & 45.9          \\
                                   \hline
                                   {VLM} &{ViT-B/32}  & {CLIP~\citep{clip}} & {151.3} & {58.0} & {43.6} & {\textbf{57.8}} & {43.6} & {45.0} & {42.3} & {43.6} & {73.1} & {53.3} & {\textbf{76.6}} & {60.1} & {53.5} & {\underline{58.8}} & {56.0}\\
                                   \hline
\multirow{5}{*}{DA}                & \multirow{2}{*}{ResNet101}                                               & DANN~\citep{dann}                             & 42.5                                &33.9 & 47.7 & 17.0 & 20.6 & 57.3 & 24.1 & 33.9 & 24.3 & 28.3 & 16.4 & 14.2 & 39.8 & 24.7 & 30.5               \\
                                   &                                               & DA-MAIC~\citep{da-maic}                          & 44.9                                & 25.8          & 28.0          & 3.1           & 5.1           & \underline{{92.6}} & 2.9           & 5.6           & 32.6          & 48.2          & 12.2          & 14.4          & 50.2          & 31.9          & 39.0          \\
                                   \cline{2-18}
                                   & \multirow{3}{*}{TResNet-M}                             & DANN~\citep{dann}                             & 29.4                                & 40.0          & \underline{82.4} & 17.2          & 27.4          & \textbf{93.8} & 17.5          & 29.5          & 67.0 & 76.8 & 23.3          & 32.6          & \textbf{93.1} & 20.4          & 33.4          \\
                                   &                                                        & DA-MAIC~\citep{da-maic}                          & 31.8                                & \underline{{62.3}} & {{77.4}} & {{42.6}} & \underline{{51.6}} & 83.1          & \textbf{51.0} & \textbf{63.2} & {{74.3}} & \underline{84.5} & {{53.9}} & \underline{{63.0}} & 83.7          & {{57.7}} & \underline{{68.3}} \\
                                   & & {{DDA-MLIC} }~\citep{dda-mlic}   & {29.4}            & {61.4}& {\textbf{84.7}} & {28.1}& {39.4}& {90.9}& {33.3}& {48.8}& {\textbf{77.0}}& {\textbf{86.9}}& {29.3}& {38.2}& {88.4}& {35.3}& {50.4}\\
                                   \cline{3-18}
                                   &                                                        & \textbf{DA-AGCN (Ours)}          & 36.6                                & \textbf{62.9}              & {75.8}              & \underline{46.8 }             & \textbf{54.0}              & 80.0              & \underline{{50.4}}              & \underline{{61.8}}              & {\underline{75.8}} & {70.9} & {\underline{71.8}} & {\textbf{69.8}} & {66.1} & {\textbf{75.6}} & {\textbf{70.5}}          \\
                                   \hline
\end{tabular}}
\end{table*}

\subsubsection{Quantitative analysis}
\paragraph{Comparison with state-of-the-art methods in terms of mAP and model size}
The proposed domain adaptation approach for multi-label image classification is compared with state-of-the-art methods. The same metrics described in Section~\ref{sec:exp_quant_agcn} are used, including mAP, CP, CR, CF1, OP, OR, and OF1. In addition to the four protocols followed in DA-MAIC~\citep{da-maic}, i.e., AID $\rightarrow$ UCM, UCM $\rightarrow$ AID, AID $\rightarrow$ DFC, and UCM $\rightarrow$ DFC, two more combinations VOC $\rightarrow$ Clipart and Clipart $\rightarrow$ VOC are provided. Two categories of methods are considered in our evaluation: (1) the conventional Multi-Label Image Classification (MLIC); and the Domain Adaptation-based (DA) methods that aim at explicitly reducing the gap between source and target datasets.

\begin{figure*}[!t]
  \centering
\begin{minipage}[b]{.49\linewidth}
  \centerline{\includegraphics[width=\linewidth]{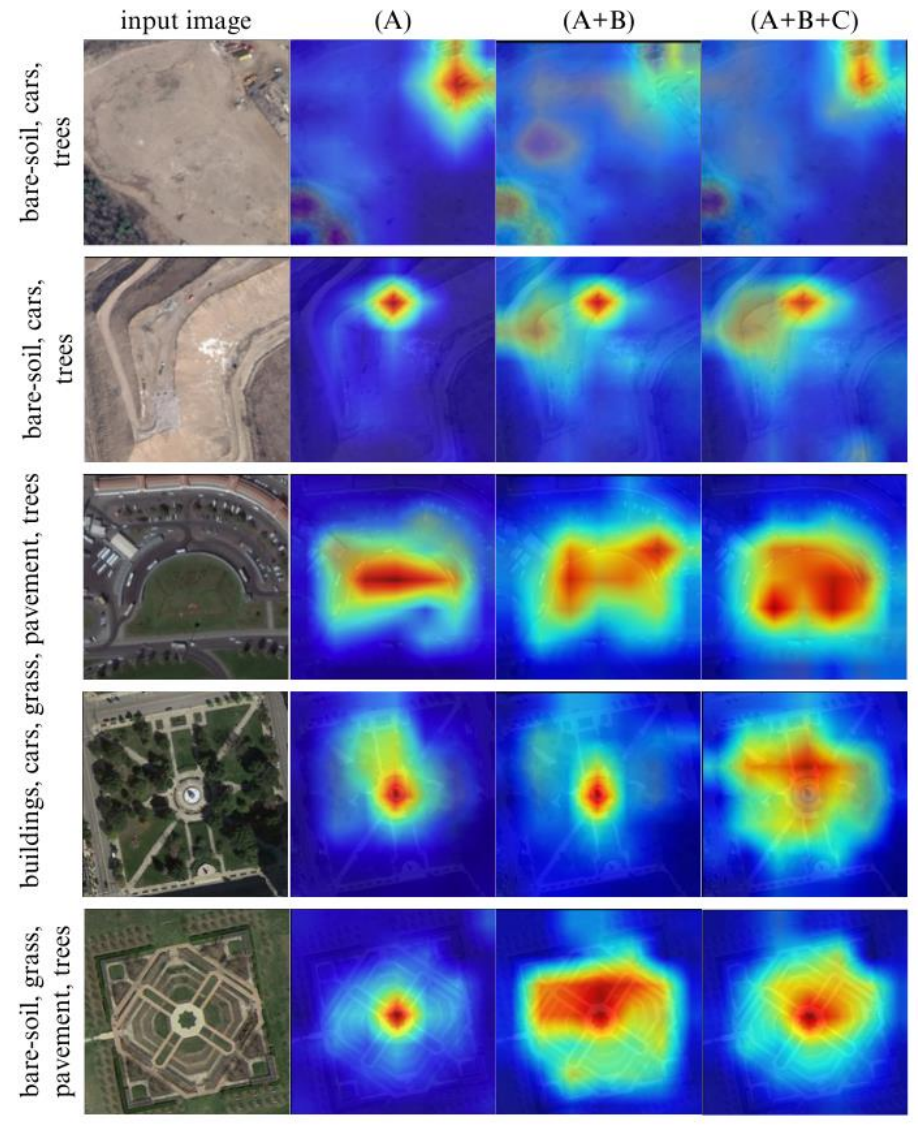}}
  \centerline{(a) UCM $\rightarrow$ AID}\medskip
\end{minipage}
\begin{minipage}[b]{.49\linewidth}
  \centerline{\includegraphics[width=\linewidth]{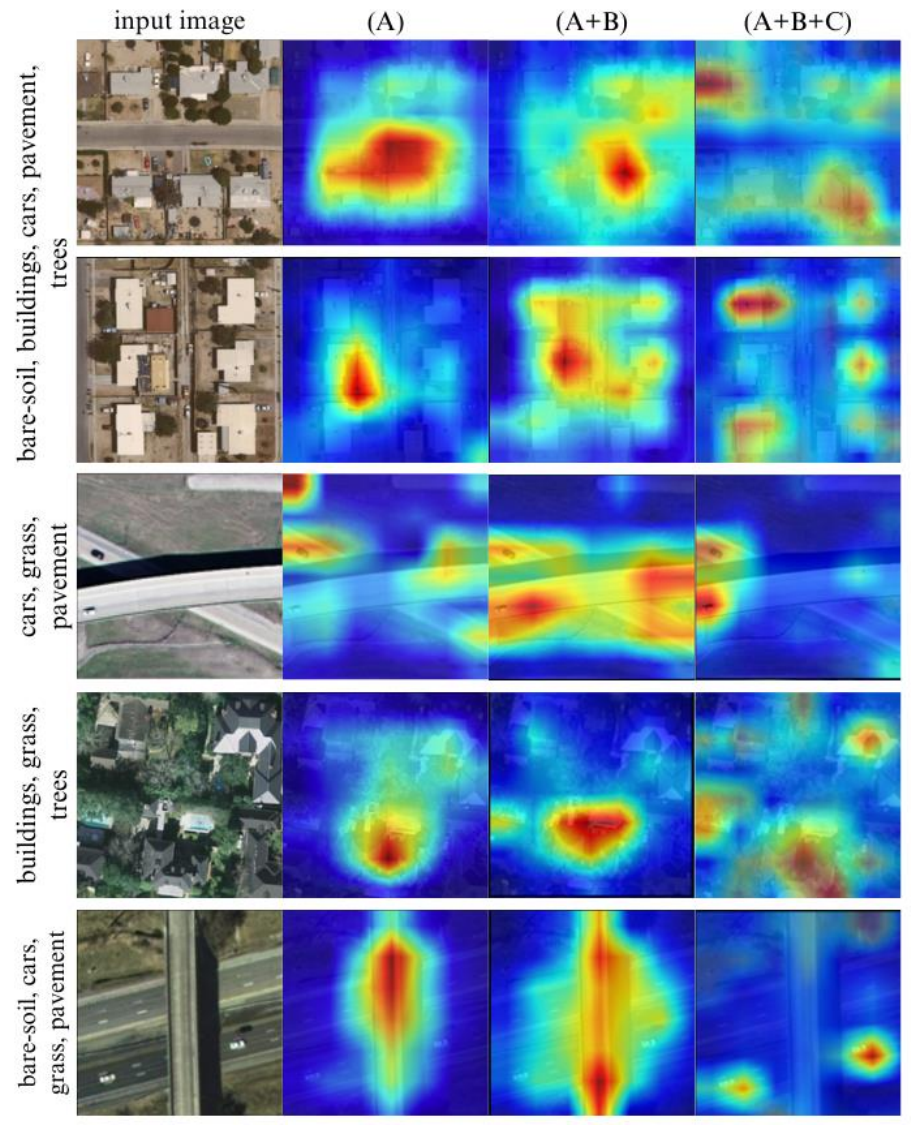}}
  \centerline{(b) AID $\rightarrow$ UCM}\medskip
\end{minipage}
\hfill
\caption{Grad-CAM visualization of the {multi-label} predictions using the proposed DA-AGCN a) UCM $\rightarrow$ AID, and b) AID $\rightarrow$ UCM setting using only $\mathbf A$, then  $\mathbf{A}$ and $\mathbf{B}$, and finally $\mathbf{A}$, $\mathbf{B}$ and  $\mathbf{C}$.}
\label{fig:qual2_aid_ucm}
\end{figure*}
Table~\ref{table:aid_ucm_results_both} reports the results obtained considering AID and UCM datasets. Two different settings are followed: AID $\rightarrow$ UCM, where AID is the source dataset and UCM is the target dataset, and UCM $\rightarrow$ AID, where UCM is the source dataset and AID is the target dataset. 
{It can be clearly seen, in both types of DA settings, that the proposed DA-AGCN outperforms existing state-of-the-art for UCM $\rightarrow$ AID. However, we are the second best in terms of mAP for AID $\rightarrow$ UCM, where the recent discriminator-free DDA-MLIC~\citep{dda-mlic} achieves a higher mAP. }
On the other hand, an improvement of mAP by 10.65\% and 4.59\% has been respectively recorded for AID $\rightarrow$ UCM and UCM $\rightarrow$ AID as compared with the DA-MAIC~\citep{da-maic} baseline. This suggests that learning the graph topology in an end-to-end manner helps improve the performance in the presence of a domain shift. {Moreover, the proposed method outperforms existing state-of-the-art VLM method by more than 17\% in terms of mAP despite having approximately five times the lesser number of model parameters.}

In Table~\ref{table:dfc_results_comb}, DA-AGCN is also compared with existing methods by considering the settings UCM $\rightarrow$ DFC and AID $\rightarrow$ DFC. It can be clearly noticed that DA-AGCN outperforms the existing works, {including {the VLM and} very recent DDA-MLIC} in terms of mAP and model size for both UCM $\rightarrow$ DFC and AID $\rightarrow$ DFC. More precisely, an improvement of 1.4\% and 6.2\% is achieved in terms of mAP compared to the second-best performing methods. 

Table~\ref{table:voc_results_comb} also confirms the superiority of the proposed approach as compared to other state-of-the-art techniques. However, it is to note that the improvement given by the adaptive graph remains limited {for Clipart $\rightarrow$ VOC, with a slightly lower performance than DDA-MLIC}. This might be explained by the fact that DA for MLIC datasets include a relatively low number of classes (8 to 20 categories). Hence, the interest of adaptively modeling label correlations might be not entirely visible. This demonstrates the necessity of creating benchmarks for DA under an MLIC context, including a wider number of classes.  

{In summary, our approach achieves the best performance for 4 settings over 2  and the second-best performance for the two remaining ones, ranking just after DDA-MLIC. This highlights the relevance of the proposed adaptive method under the unsupervised domain adaptation settings. In future works, it will be interesting to study the complementarity of the two best-performing approaches, namely, ours and DDA-MLIC.}
\begin{table*}[!t]
\centering
\caption{Ablation study: effect of adaptively learning $\mathbf{B}$ and $\mathbf{C}$ in the presence of a domain shift.}
\label{table:abl_da_graphs}
\resizebox{0.9\textwidth}{!}{%
\begin{tabular}{|ccc||c|c|c|c|c|c|c|}
\hline
\multicolumn{3}{|c||}{\textbf{Graphs}} &\multicolumn{6}{c|}{\textbf{mean Average Precision (mAP \%)}}\\
\hline
 \textbf{A} &\textbf{B} &\textbf{C} & \textbf{AID $\rightarrow$ UCM} & \textbf{UCM $\rightarrow$ AID}  & \textbf{AID $\rightarrow$ DFC} & \textbf{UCM $\rightarrow$ DFC} & {\textbf{VOC $\rightarrow$ Clipart}} & {\textbf{Clipart $\rightarrow$ VOC}} \\
\hline
\checkmark & &      & 33.6    & 42.8    & 42.6    & 58.3    & 53.1        & 68.3        \\
& {\checkmark} &      & {55.3}    & {50.6}    & {53.7}    & {69.7}    & {61.3}        & {\textbf{80.0}}        \\
& & {\checkmark}       & {47.0}    & {51.9}    & {56.7}    & {70.0}    & {50.9}        & {43.4}        \\
\checkmark &\checkmark &   & 45.1    & 55.4    & 51.5    & 67.9    & 52.5        & 75.3        \\
& {\checkmark}  &{\checkmark}   & {54.7}    & {52.7}    & {60.0}    & {71.8}    & {58.6}        & {76.1}        \\
{\checkmark} & &{\checkmark}     & {48.1}    & {54.5}    & {53.7}    & {71.6}    & {62.0}        & {75.6}        \\
\checkmark &\checkmark &\checkmark  & \textbf{59.0}    & \textbf{57.2}    & \textbf{65.7}    & \textbf{76.5}    & \textbf{62.9}        & 75.8 \\

\hline
\end{tabular}}
\end{table*}

\paragraph{Ablation study}
{Table~\ref{table:abl_da_block} reports the obtained results when considering each module comprised in DA-AGCN. More specifically, the first row reports the mAP score on the target dataset using a model trained on the source dataset without incorporating the learning of label correlations and without any domain adaptation. The results in the second row are obtained by incorporating the proposed adaptive graph learning strategy for modeling the label correlations (ML-AGCN). It can be clearly seen that adaptively learning the label graph topology based on the proposed graphs $\mathbf B$ and $\mathbf C$ leads to a significant performance improvement. More specifically, an mAP improvement of at least 16 to 20\% for UCM $\rightarrow$ AID, AID $\rightarrow$ UCM and UCM $\rightarrow$ DFC can be observed. Note that the results in the second row are obtained without any domain adaptation. Finally, the last row reports the results when adding an additional domain classifier to implicitly minimize the domain gap. This further improves the mAP score by an additional 2 to 6\%, compared to the ML-AGCN, across all benchmarks, by this remains marginal as compared to the enhancement induced by the graph structure adaptive learning. This might return to the fact that the domain gap in aerial datasets remains relatively small, hence benefiting more from the adaptive graph strategy than the alignment.}

{Additionally, in Table~\ref{table:abl_da_graphs}, we report the performance obtained on five datasets by incorporating different combinations of $\mathbf A$, $\mathbf{B}$ and $\mathbf C$. It can be noted in the first row that the original adjacency matrix $\mathbf A$ alone cannot effectively model the label relationship, hence leading to the lowest mAP score across almost all benchmarks. The next two rows of the table further showcase that using either $\mathbf B$ or $\mathbf C$ provides a significant performance improvement when compared to only $\mathbf A$, across all datasets except VOC and Clipart where $\mathbf C$ alone does not yield superior performance as compared to $\mathbf A$. This not surprising since $\mathbf C$ aims to preserve the node feature similarity, and is not expected to model the label correlations. Furthermore, the next three rows show the variation in performance when coupling two to three matrices. For instance, combining either $\mathbf B$ or $\mathbf{C}$ with the original adjacency matrix $\mathbf{A}$ always provides a higher mAP when compared to only $\mathbf{A}$. Finally, the last row confirms the superiority of the proposed combination $\mathbf{A}, \mathbf{B}$ and $\mathbf{C}$ as the highest mAP score across all benchmarks is reached except for Clipart $\rightarrow$ VOC. A possible explanation to this exception is the synthetic nature of the Clipart dataset, where the label co-occurrence is well-thought and might be representative enough.}


  {Furthermore, the importance of preserving the node feature similarity is demonstrated in the last row, where a notable improvement in mAP is observed as compared to the $\mathbf A + \mathbf B$ setting. As compared to the single-domain setup, the improvement resulting from $\mathbf C$ is more significant (see Table~\ref{table:ablation_mlic}, where the contribution of $\mathbf C$ is relatively marginal). This might be explained by the fact that in a cross-domain setting, an adversarial training is adopted for enforcing the generation of domain-invariant features. This min-max game known for its instability, might amplify the node feature dissimilarity phenomenon.}
 

\begin{table}[!t]
\caption{Ablation study: impact of using an adversarial strategy and learning an adaptive graph topology.}
\label{table:abl_da_block}
\resizebox{0.48\textwidth}{!}{%
\begin{tabular}{|l|c|c|c|c|}
\hline
\textbf{Methods/Dataset   (mAP)} & \textbf{UCM $\rightarrow$ AID} & \textbf{AID $\rightarrow$ UCM} & \textbf{AID $\rightarrow$ DFC} & \textbf{UCM $\rightarrow$ DFC} \\
\hline
TResNet only            & 34.98     & 37.03     & 62.36         & 54.77\\
TResNet + AGCN          & 53.62 \textbf{(+18.64)}     & 53.08 \textbf{(+16.05)}     & 58.44 (-3.92)       & 74.4 \textbf{(+19.68)} \\
TResNet + AGCN + DC     & 55.79 \textbf{(+2.17)}    & 55.37 \textbf{(+2.29)}    & 65.11 \textbf{(+6.67)}       & 76.49 \textbf{(+2.04)} \\
\hline
\end{tabular}}
\end{table}


\begin{table}[!t]
\centering
\caption{{Sensitivity analysis: performance of the proposed DA-AGCN using different values of the hyper-parameter ($\lambda$), defined in Eq.~\eqref{eq:lambda}.} }
\label{table:sensitivity}
\resizebox{0.48\textwidth}{!}{%
\begin{tabular}{|c|c|c|c|c|c|c|}
\hline
\multirow{2}{*}{lambda $(\lambda)$} & \multicolumn{6}{c|}{mean Average Precision   (mAP \%)}                                                                                                                                     \\
\cline{2-7}
                        & \multicolumn{1}{l|}{AID$\rightarrow$UCM} & \multicolumn{1}{l|}{UCM$\rightarrow$AID} & \multicolumn{1}{l|}{AID$\rightarrow$DFC} & \multicolumn{1}{l|}{UCM$\rightarrow$DFC} & \multicolumn{1}{l|}{VOC$\rightarrow$Clipart} & \multicolumn{1}{l|}{Clipart$\rightarrow$VOC} \\
                        \hline
0.1                     & 52.2                        & 52.5                        & 56.5                        & 68.4                        & 51.2                            & 73.9                            \\
0.2                     & 53.3                        & 54.9                        & 53.1                        & 70.6                        & 52.8                            & \textbf{75.8}                   \\
0.3                     & 53.5                        & 52.0                        & 58.2                        & \textbf{72.8}               & 52.6                            & 71.6                            \\
0.4                     & 51.7                        & \textbf{57.2}               & 57.0                        & 70.5                        & 50.0                            & 74.2                            \\
0.5                     & 51.6                        & 56.7                        & 55.1                        & 71.1                        & 52.1                            & 73.9                            \\
0.6                     & \textbf{57.0}               & 56.3                        & \textbf{58.9}               & 69.8                        & 52.6                            & 72.7                            \\
0.7                     & 54.7                        & 52.2                        & 56.9                        & 71.1                        & 52.7                            & 72.5                            \\
0.8                     & 50.4                        & 53.1                        & 55.4                        & 69.3                        & 52.5                            & 73.8                            \\
0.9                     & 53.5                        & 54.2                        & 55.7                        & 72.1                        & \textbf{53.0}                   & 75.3                            \\
1.0                     & 53.4                        & 53.9                        & 56.4                        & 70.7                        & 52.5                            & 74.0                           \\
\hline
\end{tabular}}
\end{table}


\paragraph{{Hyper-parameter analysis}}
{In Table~\ref{table:sensitivity}, we report the mean Average Precision (mAP) on the six considered benchmarks when varying the hyper-parameter ($\lambda$), introduced in Eq~\eqref{eq:lambda}. Specifically, we vary its value from $0.1$ to $1.0$.  It can be noted that the obtained results are relatively stable with a variation in maP of more or less 5\%.}

\subsubsection{Qualitative analysis}
Grad-CAM~\citep{grad-cam} visualization in the presence of a domain shift can be seen in Fig.~\ref{fig:qual2_aid_ucm}. While Fig.~\ref{fig:qual2_aid_ucm} (a) demonstrate these results for UCM $\rightarrow$ AID, Fig.~\ref{fig:qual2_aid_ucm} (b) showcases the visualizations for AID $\rightarrow$ UCM. The left-most column is the input image, and the next three consecutive columns represent the Grad-CAM visualization of the image using a model trained with; $\mathbf{A}$, $\mathbf{A}$ and $\mathbf{B}$, and $\mathbf{A}$, $\mathbf{B}$, $\mathbf{C}$ respectively. It can be clearly seen that adaptively learning the graph topology $\mathbf{B}$ and $\mathbf{C}$ helps activate the most relevant areas of interest, leading to better classification performance.

\subsubsection{Failure cases}
\begin{figure}[!t]
  \centering
  \centerline{\includegraphics[width=1.0\linewidth]{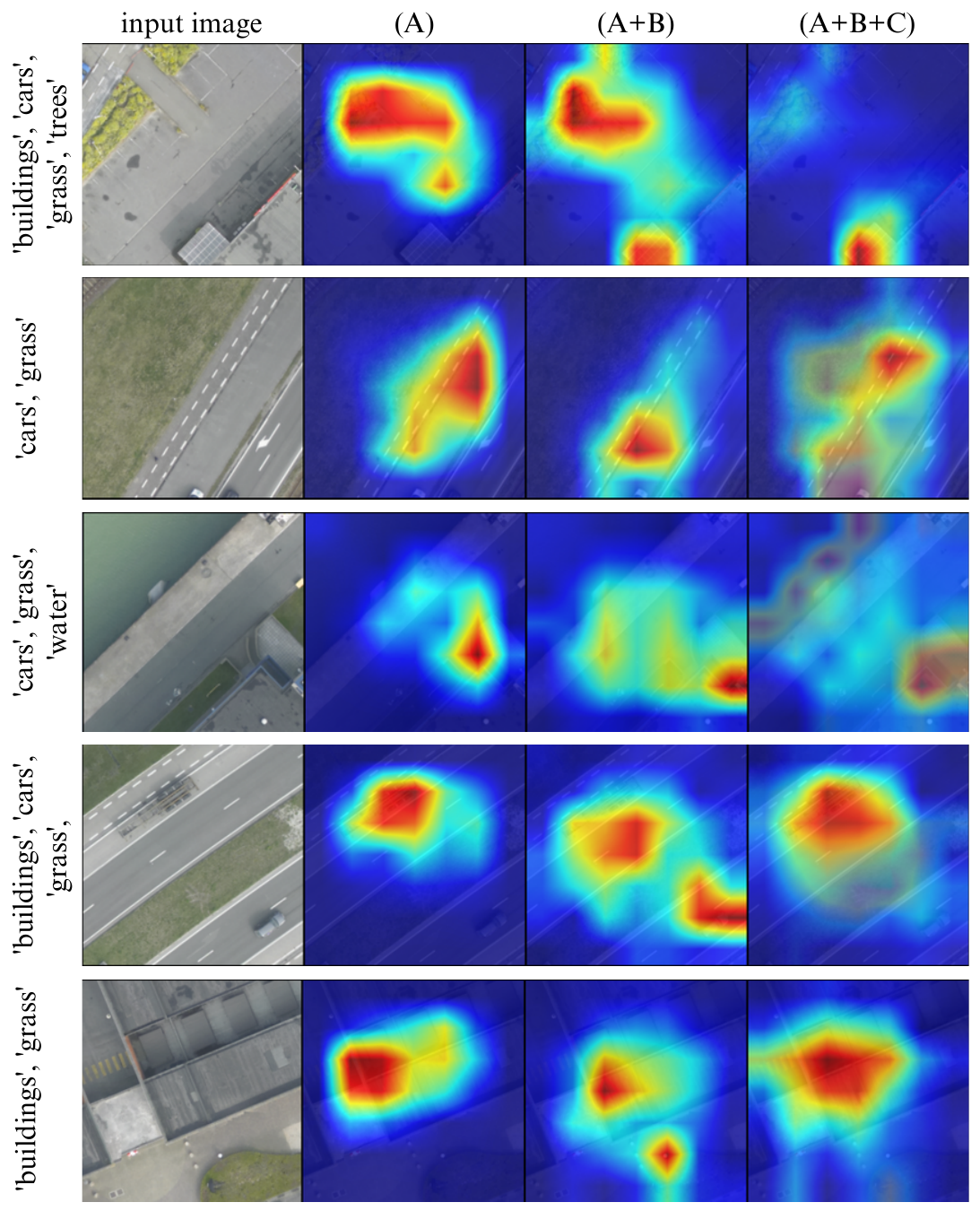}}
\caption{Failure cases: Grad-CAM visualization for AID $\rightarrow$ DFC {with ground truth multi-labels} showcasing performance degradation with adaptive graphs.}
\label{fig:failure}
\end{figure}

In Fig.~\ref{fig:failure}, some failure cases are presented using the Grad-CAM visualization. {The first column shows the input image with ground truth labels while the subsequent three columns are the activated Grad-CAMs with $\mathbf{A}$, $\mathbf{A}$ and $\mathbf{B}$ and $\mathbf{A}, \mathbf{B}$ and $\mathbf{C}$, respectively.} It can be noticed that when the targeted object occupies most of the image, the use of $\mathbf B$ and $\mathbf C$ makes the model confuse the object with the background. A good illustration of this can be seen in the second row of Fig.~\ref{fig:failure}. The \textit{car} is confused with the \textit{grass}. In future work, modeling the object occupancy will be investigated for handling these failure cases.

\section{Conclusion}
\label{sec:conclusion}
Existing graph-based methods have shown great performances for multi-label image classification in the context of both single-domain and cross-domain. However, these methods mostly fix the graph topology heuristically while discarding edges with rare co-occurrences. Furthermore, it has been demonstrated in~\citep{nodesim} that successive GCN aggregations tend to destroy the initial feature similarity.
Hence, as a solution, an adaptive strategy for learning the graph in an end-to-end manner is proposed. In particular, attention-based and similarity-preserving mechanisms are adopted. The proposed framework for multi-label classification in a single domain is then extended to multiple domains. For that purpose, an adversarial domain adaptation strategy is employed. The results performed for both single and cross-domain support the effectiveness of the proposed method in terms of model performance and size as compared to recent state-of-the-art methods. 
{The current work is restricted to scenarios where the source and target data have shared object categories. Nevertheless, in future works, we intend to investigate a more challenging problem, namely, open set domain adaptation where only a few categories of labels are shared between source and target data.}

\section*{Acknowledgments}
This research was partially funded by the Luxembourg National Research Fund (FNR), grant references BRIDGES2020/IS/14755859/MEET-A/Aouada. For the purpose of open access, and in fulfillment of the obligations arising from the grant agreement, the author has applied a Creative Commons Attribution 4.0 International (CC BY 4.0) license to any Author Accepted Manuscript version arising from this submission.





\bibliographystyle{model2-names}
\bibliography{refs}



\end{document}